\documentclass[
]{ceurart}

\usepackage{listings}
\usepackage{subfig}
\usepackage{hyperref} 

\begin{document}

\copyrightyear{2026}
\copyrightclause{Copyright for this paper by its authors.
Use permitted under Creative Commons License Attribution 4.0 International (CC BY 4.0).}
\conference{EXPLIMED 2026 - Third Workshop on Explainable Artificial Intelligence for the medical domain - 15-17 August 2026, Bremen, Germany}

\title{TRACE: A Concept Bottleneck Model for Longitudinal 3D Glioblastoma Response Assessment}

\author[1]{Alia Tarek}[%
orcid={0009-0009-7824-4500},
]
\fnmark[1]
\address[1]{Department of Biomedical and Healthcare Data Engineering, Faculty of Engineering, Cairo University, Giza, Egypt}

\author[1]{Hamsa Saber}[%
orcid={0009-0005-9866-3375},
]
\fnmark[1]

\author[1]{Hamza Elghonemy}[%
orcid={0009-0007-5277-9545}
]
\fnmark[1]

\author[1]{Youssef Afify}[%
orcid={0009-0001-9486-0997},
]
\fnmark[1]

\author[1]{Tamer Basha}[%
orcid={0000-0003-4431-8646},
email={tamer.basha@eng1.cu.edu.eg},
]

\author[2]{Omair Shahzad Bhatti}[%
orcid={0000-0001-7983-2384},
email={omair_shahzad.bhatti@dfki.de},
]
\address[2]{German Research Center for Artificial Intelligence (DFKI), Saarbrücken, Germany}

\author[2,3]{Abdulrahman M. Selim}[%
orcid={0000-0002-4984-6686},
email={abdulrahman.mohamed@dfki.de},
]
\cormark[1]
\address[3]{University of Oldenburg, Oldenburg, Germany}

\author[2]{Hasan Md Tusfiqur Alam}[%
orcid={0000-0003-1479-7690},
email={hasan.alam@dfki.de},
]
\cormark[1]

\author[2,3]{Daniel Sonntag}[%
orcid=0000-0002-8857-8709,
email={daniel.sonntag@dfki.de},
]

\fntext[1]{These authors contributed equally and are listed alphabetically.}
\cortext[1]{Corresponding authors.}

\begin{abstract}
  Longitudinal glioblastoma response assessment requires comparing subtle tumor changes across MRI time points according to structured clinical criteria such as RANO. However, most deep learning approaches predict response labels directly from imaging features, which limits inspection and correction. Therefore, we introduce \textsc{Trace}, an exploratory RANO 2.0-aligned concept bottleneck model for interpretable 4-class response classification evaluated on the LUMIERE dataset. \textsc{Trace} processes paired baseline and follow-up multimodal MRI with a shared 3D vision encoder, predicts clinically meaningful root measurements from the tumor, computes downstream RANO-derived concepts deterministically, and adds scan interval and new-lesion information as passthrough concepts. This frames response assessment as structured concept reasoning rather than direct image-to-label prediction.
  Using 5-fold patient-wise cross-validation, \textsc{Trace} achieves a 4-class macro F1 of $0.4769 \pm 0.1229$ and a binary progression-versus-non-progression macro F1 of $0.7085 \pm 0.0935$. It improves over a concept bottleneck baseline and remains within the range of published non-interpretable deep learning approaches. Ablations show that the expert RANO graph and intervention-consistency training are both important, and intervention experiments show that correcting concepts can improve downstream predictions. These findings support structured concept bottlenecks as a transparent direction for longitudinal GBM response assessment, while motivating larger protocol-aligned datasets for rare-class performance and external validation. Our codebase is publicly available online on \href{https://github.com/yusufafify/TRACE-Temporal-Causal-Reasoning-with-Acyclic-Concept-Explanation}{GitHub}.
\end{abstract}


\maketitle

\section{Introduction}\label{sec:introduction}
Glioblastoma (GBM), a World Health Organization (WHO) grade 4 astrocytoma, is the most common and aggressive primary malignant brain tumor in adults, and accounts for about half of all gliomas \cite{thakkar2014glioblastoma}.
Because disease progression is nearly universal, accurate detection of tumor recurrence on longitudinal Magnetic Resonance Imaging (MRI) is essential for timely treatment decisions. 
Clinicians commonly assess the treatment response using the Response Assessment in Neuro-Oncology (RANO) criteria. Although the updated RANO 2.0 criteria improve assessment by requiring confirmatory scans within 12 weeks after radiotherapy \cite{wen2023rano}, conventional MRI alone has limited ability to distinguish treatment-related changes, such as pseudoprogression, from true tumor progression.
This limitation highlights the need for more objective and automated decision-support methods for GBM treatment response assessment.

These challenges have motivated an increased interest in Artificial Intelligence (AI) methods for longitudinal MRI analysis.
AI-based approaches have been applied across the full spectrum of GBM care, from initial diagnosis through treatment planning and longitudinal monitoring. 
Among these tasks, treatment response assessment is especially challenging because it requires detecting small changes over time rather than findings from a single scan; this is difficult because adjacent time points can look highly similar, which can cause standard models to miss subtle biomarkers of disease evolution \cite{zhang2025glomiaprogeneralizablelongitudinalmedical}.
Deep learning approaches have shown promising results; for example, a 3D DenseNet model achieved strong performance in distinguishing pseudoprogression from true progression \cite{moassefi2022deep}. 
However, most AI models for GBM still face critical challenges in clinical use, such as data heterogeneity, small sample sizes, and limited interpretability \cite{khalighi2024ai, roncevic2025ai, ghadimi2024segmentation}.

Interpretability is particularly important because treatment response is assessed according to explicit clinical criteria.
Deep learning models often do not expose the reasoning behind their predictions, a limitation that has been widely discussed in oncology AI and in brain tumor MRI analysis \cite{hagenbuchner2020blackbox,gulum2021explainable, charaabi2023xai}. 
A common response has been to apply post-hoc explainability techniques to black-box architectures \cite{desai2025xai}, such as Gradient-weighted Class Activation Mapping (Grad-CAM) \cite{selvaraju_grad-cam_2017}. 
However, such explanations are not always reliable or clinically meaningful.
The limitations of post-hoc explainability have motivated a shift toward inherently interpretable models. 
Concept Bottleneck Models (CBMs)~\cite{koh2020concept, alam_towards_cbm} offer an inherently interpretable alternative by predicting explicit intermediate concepts before the final label. This is useful in medical imaging because clinicians can inspect concept predictions and, when needed, correct them at test time so that corrections propagate to the output~\cite{shin2023closerlookinterventionprocedure, alam_cbm_rag}.
Despite growing interest in CBMs across medical imaging tasks, e.g., using causal CBMs \cite{defelice2026causallyreliableconceptbottleneck} for chest x-ray interpretation that integrate radiologists’ knowledge and demonstrate improved calibration, to the best of our knowledge, no prior work has applied CBMs to longitudinal brain-tumor progression or explicitly encoded temporal and causal dependencies in a concept graph.

In this paper, we introduce \textsc{Trace} (\textbf{T}emporal Causal \textbf{R}easoning with \textbf{A}cyclic \textbf{C}oncept \textbf{E}xplanation), a RANO-aligned temporal CBM for interpretable GBM treatment response classification.
\textsc{Trace} processes paired baseline and follow-up multimodal MRI with a shared-weight siamese 3D encoder, predicts clinically grounded tumor measurements as root concepts, and computes downstream RANO-derived concepts through deterministic nodes. This frames longitudinal response assessment as a structured concept reasoning problem rather than a direct image-to-label prediction task.

Our contributions are as follows:
\begin{itemize}\setlength\itemsep{0.2em}
    \item We propose \textsc{Trace}, a RANO-aligned CBM architecture for longitudinal GBM response assessment. To our knowledge, this is the first CBM designed for RANO-based longitudinal GBM response classification. Unlike standard CBMs that treat concepts as a flat set of predictors, \textsc{Trace} represents clinically defined dependencies as a directed acyclic graph over tumor measurements, percentage changes, threshold flags, passthrough metadata, and the final response label.

    \item We encode deterministic RANO reasoning inside the bottleneck. Root tumor measurements are estimated from paired MRI, while downstream RANO quantities are recomputed from their parent concepts. This keeps derived concepts mutually consistent, constrains the task head to human-interpretable bottleneck variables, and allows concept-level corrections to propagate through the response pathway.

    \item We evaluate \textsc{Trace} on the publicly available LUMIERE dataset~\cite{suter2022lumiere} using 5-fold patient-wise cross-validation. \textsc{Trace} achieves a 4-class macro F1 of $0.4769 \pm 0.1229$, improves over a data-efficient CBM baseline, and remains within the range of published non-interpretable deep learning baselines. We further use ablations, concept-effect analysis, and intervention experiments to examine which parts of the structured bottleneck contribute to performance and where the model remains limited.
\end{itemize}

Overall, this work shows that explicitly encoding temporal and clinically specified dependencies in a CBM can support more interpretable GBM response assessment. We present \textsc{Trace} as an exploratory step under the practical constraints of small and imbalanced clinical datasets, and as a design pattern that may extend to other longitudinal medical imaging tasks where decisions depend on explicit clinical criteria, temporal change, and human-verifiable intermediate measurements.
Our codebase is publicly available online on \href{https://github.com/yusufafify/TRACE-Temporal-Causal-Reasoning-with-Acyclic-Concept-Explanation}{https://github.com/yusufafify/TRACE-Temporal-Causal-Reasoning-with-Acyclic-Concept-Explanation}.

\section{Related Work}
Prior studies on automated GBM MRI analysis have mostly focused on tumor segmentation and volumetric assessment instead of direct treatment response classification. 
Tools such as BraTumIA~\cite{abu_khalaf2021} and HD-GLIO-AUTO~\cite{kickingereder2019automated,Isensee_2020,isensee_automated_2019} can automatically segment tumors and estimate quantitative imaging measures from multimodal MRI. However, these systems still rely on clinicians to interpret the outputs and apply the RANO criteria for response assessment \cite{suter2023evaluating}.

Recent work has begun to predict RANO response categories directly from longitudinal MRI using multi-timepoint CNNs \cite{matoso2025deeplearningapproachclassifying}, radiomics \cite{amato2025integrating}, or hybrid deep learning and radiomics models~\cite{tikhonov2025predicting}. 
These methods provide important performance references, but they predict labels directly and do not address the need for clinically interpretable reasoning under the RANO framework.
The difficulty is that RANO assessment depends on explicit clinical evidence, including spatial localization of new lesions, measurement of enhancing tumor components, and comparison across imaging time points. Models that predict only the final response label do not make this reasoning visible, making their outputs difficult to interpret and verify in practice. This limits their usefulness in clinical settings, where decision-support systems are expected to provide reasoning that clinicians can understand and assess \cite{gulum2021explainable}. 

\subsection{Concept Bottleneck Models for Structured and Interpretable Prediction}
CBMs provide a natural way to connect image-based prediction with structured clinical reasoning. Instead of mapping an input directly to a label, CBMs first predict explicit intermediate concepts and then use these concepts to derive the final decision \cite{koh2020concept}. This makes them well-suited to RANO-based response assessment, where the final class depends on clinically meaningful intermediate criteria rather than on image appearance alone.

Beyond static prediction, recent work has begun to extend concept bottlenecks to time-dependent data. X-CHAR~\cite{jeyakumar_x-char_2023} models complex human activity recognition as sequences of human-understandable concepts, while MoTIF~\cite{knab_concepts_2026} uses per-concept temporal self-attention to model concept dynamics in video classification. These works show that temporal concept modeling is an emerging direction, but the studied settings remain far from longitudinal medical response assessment, where predictions must follow explicit clinical criteria across patient time points.

Recent work has also extended CBMs in directions relevant to medical imaging. Some studies use large language models (LLMs) or vision-language models (VLMs) to define or align concepts automatically, reducing the need for fully manual concept annotation \cite{bai2024m3d}. Others have shown that CBMs can improve robustness and reduce reliance on spurious correlations compared with standard visual encoders \cite{yan2023robustinterpretablemedicalimage}. 
In breast ultrasound, \citet{bunnell2024learningclinicallyrelevantconceptbottleneck} showed that clinician intervention on predicted concepts can improve downstream classification, using BI-RADS features as the concept set~\cite{magny2023}.

At the same time, standard CBMs still have practical limitations. They often require concept annotations, which are expensive to obtain, especially in specialized domains such as neuro-oncology. They can also require larger datasets and may still learn undesirable concept-label associations if the concept layer is not designed carefully \cite{kim2024constructingconceptbasedmodelsmitigate,fokkema2025sampleefficientlearningconceptstheoretical}. 
Several extensions have been proposed to address these issues; for example, Label-Free CBMs reduce the burden of manual annotation by using multimodal foundation models to discover or align concepts automatically \cite{wu2025mmcbm,oikarinen2023labelfreeconceptbottleneckmodels}; Data-Efficient CBMs (DCBM) aim to improve concept learning when only limited training data are available \cite{prasse2025dcbmdataefficientvisualconcept}; and Causal CBMs (C$^2$BM) further model structured dependencies between concepts and predictions to reduce reliance on shortcut correlations \cite{defelice2026causallyreliableconceptbottleneck}.

Within this broader literature, two CBM variants are especially relevant to our setting. \textbf{DCBMs} are designed for data-sparse scenarios, making them a natural point of comparison when training data are limited. 
In addition, \textbf{C$^2$BM} \cite{defelice2026causallyreliableconceptbottleneck} is particularly relevant because it combines concept-based prediction with an explicit directed acyclic graph (DAG) over concepts and the task label, which constrains information flow to predefined dependencies and supports propagation of concept-level interventions through the reasoning chain. 
For longitudinal GBM response assessment, these two directions address complementary needs, i.e., data efficiency and structured clinical reasoning. We compare these variants to identify which strategy better balances predictive accuracy and clinical interpretability for longitudinal GBM response assessment.

\subsubsection{Directed Acyclic Graphs for Structured Clinical Reasoning}

RANO assessment follows an explicit clinical decision process in which measurable findings, lesion status, and temporal evidence are combined to determine the final response category. DAGs provide a natural way to represent these dependencies in an explicit and interpretable form. In clinical and epidemiological research, DAGs are used to encode expert knowledge, make assumed causal pathways explicit, and support adjustment-variable selection in observational studies \cite{Shrier2008,Evans2012}. More recently, DAG-based reasoning has also been discussed in clinical prediction modeling, where the goal is to ensure that models reflect clinically meaningful structure rather than relying only on correlations in the data \cite{Piccininni2020}. This perspective is relevant here because RANO is already organized around structured dependencies between intermediate findings and the final response label.

However, a DAG alone does not define how these intermediate findings should be predicted from imaging data. For that, a model, such as C$^2$BM, is needed which can both learn concepts from MRI and preserve the structured reasoning process that links those concepts to the final response category.

\section{Methodology}\label{sec:siamese_cbm}

For treatment response assessment, clinicians compare measurable tumor changes between longitudinal MRI scans across two time points, a baseline scan $t_0$ and a follow-up scan $t_1$.
In accordance with the RANO 2.0 criteria~\cite{wen2023rano}, we use two complementary measurements of tumor size, which are the tumor volume $V$ in cm$^3$, estimated from segmentation masks, and the Sum of Products of Diameters (SPD), which is a 2D clinical measurement in cm$^2$ defined as the product of the two largest perpendicular tumor diameters on a single imaging slice. Both quantities are measured at baseline and follow-up, resulting in $V_{t_0}$, $V_{t_1}$, $\mathrm{SPD}_{t_0}$, and $\mathrm{SPD}_{t_1}$.
Additional clinical background on SPD is provided in Appendix~\ref{app:spd_background}.

The main idea of our method is to replace black-box temporal reasoning with an explicit, RANO-aligned concept reasoning chain. Rather than mapping paired MRI scans directly to a response label, the model first predicts clinically meaningful root concepts, derives downstream RANO concepts through deterministic operations, and then predicts the final response class from this structured bottleneck. 
This results in an auditable pathway:
\begin{equation}
(x_{t_0}, x_{t_1})
\;\longrightarrow\;
\underbrace{C_r}_{\text{root concepts}}
\;\longrightarrow\;
\underbrace{C_d}_{\text{derived RANO concepts}}
\;\longrightarrow\;
\hat{Y},
\end{equation}
where $C_r$ contains image-derived tumor measurements, $C_d$ contains deterministic RANO
deltas and threshold flags, and $\hat{Y}$ is the predicted RANO response category.

This design is motivated by the fact that RANO response assessment is already concept-based in clinical practice. Clinicians do not reason directly from voxel intensities to a response label; instead, they compare tumor measurements across time, assess threshold changes, consider new lesions and interpret the result according to a structured protocol. Our model encodes this reasoning structure as a DAG, making the intermediate steps inspectable and correctable.

\subsection{RANO 2.0 Response Criteria}
\label{sec:rano_criteria}

RANO 2.0 defines four response categories: Complete Response (CR), Partial Response (PR), Stable Disease (SD), and Progressive Disease (PD) \cite{wen2023rano}. These classes are determined by measurable changes between baseline and follow-up tumors. 
We compute percentage changes as
\begin{equation}
\Delta V_{\%}
=
\frac{V_{t_1}-V_{t_0}}{V_{t_0}+\varepsilon}\times 100,
\qquad
\Delta \mathrm{SPD}_{\%}
=
\frac{\mathrm{SPD}_{t_1}-\mathrm{SPD}_{t_0}}
{\mathrm{SPD}_{t_0}+\varepsilon}\times 100,
\label{eq:delta_concepts}
\end{equation}
where $\varepsilon$ is a small constant used for numerical stability.

Table~\ref{tab:rano_criteria} summarizes the decision thresholds used in this work. A response category can be triggered by either the volume-based or SPD-based criterion. In addition, the appearance of a new lesion directly supports PD under RANO.
Volume and SPD are treated as complementary rather than redundant measurements. 
Volume captures the 3D tumor burden through both enhancing and non-enhancing tumor volumes (i.e., complementary tumor regions visible across different MRI sequences), while SPD retains compatibility with established clinical response criteria and historical neuro-oncology trials. Including both measurements allows the model to follow the full clinical RANO reasoning chain rather than relying on a single proxy for disease extent.

\begin{table}
    \centering
    \caption{RANO 2.0 response categories and decision thresholds. A measurability floor is applied:
    $V_{t_0} \geq 0.5\,\mathrm{cm}^3$ for volume-based criteria and
    $\mathrm{SPD}_{t_0} \geq 0.01\,\mathrm{cm}^2$ for SPD-based criteria.}
    \label{tab:rano_criteria}
    \begin{tabular}{lll}
        \toprule
        \textbf{Response} & \textbf{Volume criterion} & \textbf{SPD criterion} \\
        \midrule
        Complete Response (CR)   & No measurable tumor        & No measurable tumor \\
        Partial Response (PR)    & $\Delta V_{\%} \leq -65\%$       & $\Delta\mathrm{SPD}_{\%} \leq -50\%$ \\
        Stable Disease (SD)      & $-65\% < \Delta V_{\%} < +40\%$  & $-50\% < \Delta\mathrm{SPD}_{\%} < +25\%$ \\
        Progressive Disease (PD) & $\Delta V_{\%} \geq +40\%$       & $\Delta\mathrm{SPD}_{\%} \geq +25\%$ \\
        \bottomrule
    \end{tabular}
\end{table}

\subsection{Structured Concept Graph}
\label{sec:concept_structure}

Standard CBMs~\cite{koh2020concept} typically learn a mapping from input $x$ to concepts $C$, followed by a mapping from concepts $C$ to target label $Y$:
\begin{equation}
x \rightarrow C \rightarrow Y.
\end{equation}
However, RANO concepts are not statistically independent attributes. Percentage changes are mathematical functions of baseline and follow-up measurements, and threshold flags are deterministic functions of the percentage changes. Treating these variables as independent concepts can create mutually inconsistent concept states and can weaken the reliability of test-time intervention.

Therefore, we encode the RANO reasoning process as a DAG
\begin{equation}
\mathcal{G}=(\mathcal{V},\mathcal{E}),
\end{equation}
where each node in $\mathcal{V}$ corresponds to a concept or the task label, and each edge in $\mathcal{E}$ represents a clinically defined dependency. 
The graph contains four types of nodes:
\begin{itemize}
    \item \textbf{Root concepts} $C_r$: image-derived tumor measurements predicted from MRI and
    segmentation inputs.
    \item \textbf{Derived concepts} $C_d$: deterministic RANO quantities computed from parent concepts.
    \item \textbf{Passthrough concepts} $M$: metadata or precomputed clinical indicators injected directly
    into the graph.
    \item \textbf{Task node} $Y$: the final 4-class RANO response prediction.
\end{itemize}

The complete concept set is shown in Table~\ref{tab:concepts}. Unlike causal CBM settings where the concept graph must be learned or inferred from data~\cite{defelice2026causallyreliableconceptbottleneck}, our graph is specified directly from the clinical RANO protocol. This is a key advantage of the task, i.e., the relevant dependencies are known a priori and can be encoded as architectural constraints.

\begin{table}[t]
    \centering
    \caption{Concepts in the proposed RANO-aligned concept graph. Root concepts are predicted from image features. Derived concepts are computed deterministically from their parents. Passthrough concepts are injected directly as metadata or precomputed clinical indicators.}
    \label{tab:concepts}
    \begin{tabular}{llp{2.2cm}p{6.6cm}}
        \toprule
        \textbf{Concept} & \textbf{Symbol} & \textbf{Type} & \textbf{Derivation / Source} \\
        \midrule
        Baseline volume  & $V_{t_0}$ & \multirow{4}{*}{Root} & \multirow{4}{*}{Predicted from $\mathbf{X}_{\mathrm{enc}}$ via a ConceptBlock} \\
        Follow-up volume & $V_{t_1}$ & & \\
        Baseline SPD & $\mathrm{SPD}_{t_0}$ & & \\
        Follow-up SPD & $\mathrm{SPD}_{t_1}$ & & \\
        \midrule
        Volume change & $\Delta V_{\%}$ & \multirow{6}{*}{Derived} &
        $(V_{t_1}-V_{t_0})/(V_{t_0}+\varepsilon)\times 100$ \\
        SPD change & $\Delta \mathrm{SPD}_{\%}$ & &
        $(\mathrm{SPD}_{t_1}-\mathrm{SPD}_{t_0})/(\mathrm{SPD}_{t_0}+\varepsilon)\times 100$ \\
        Volume PD flag & $b_{\mathrm{PD}}^{\mathrm{vol}}$ & &
        $\mathbf{1}[\Delta V_{\%}\geq 40\% \wedge V_{t_0}\geq 0.5]$ \\
        Volume PR flag & $b_{\mathrm{PR}}^{\mathrm{vol}}$ & &
        $\mathbf{1}[\Delta V_{\%}\leq -65\% \wedge V_{t_0}\geq 0.5]$ \\
        SPD PD flag & $b_{\mathrm{PD}}^{\mathrm{SPD}}$ & &
        $\mathbf{1}[\Delta \mathrm{SPD}_{\%}\geq 25\% \wedge \mathrm{SPD}_{t_0}\geq 0.01]$ \\
        SPD PR flag & $b_{\mathrm{PR}}^{\mathrm{SPD}}$ & &
        $\mathbf{1}[\Delta \mathrm{SPD}_{\%}\leq -50\% \wedge \mathrm{SPD}_{t_0}\geq 0.01]$ \\
        \midrule
        Scan interval & $\Delta t$ & \multirow{2}{*}{Passthrough} & Administrative metadata \\
        New-lesion flag & $b_{\mathrm{NL}}$ & & Precomputed lesion-topology indicator \\
        \bottomrule
    \end{tabular}
\end{table}

\subsection{Theoretical Properties of the Structured Bottleneck}
\label{sec:theoretical_properties}

The proposed architecture is designed to satisfy two important properties for clinical interpretability, i.e., no hidden image-feature leakage into the final classifier and deterministic propagation of concept corrections. These properties align with recent views of actionable interpretability, in which interpretable models should use variables with human-aligned semantics and structures that users can reason about \cite{barbiero_actionable_2026}.

\paragraph{Proposition 1: No-leakage bottleneck.}
Let $X$ denote the paired MRI and segmentation inputs, $C_r$ the predicted root concepts, $C_d$
the deterministic derived concepts, $M$ the passthrough metadata, and $\hat{Y}$ the final prediction.
If the task classifier receives only $(C_r,C_d,M)$ and no direct image embedding, then
\begin{equation}
\hat{Y} \perp X \mid (C_r,C_d,M).
\end{equation}
In other words, once the bottleneck variables are fixed, the final prediction has no additional access to image information.

\textbf{Interpretation.}
This property does not claim that the learned root concepts are always perfectly accurate. Rather, it states that the final classifier cannot use hidden image features outside the concept graph. This reduces the risk of information leakage, a known problem in CBMs where unintended information beyond the intended concepts can be exploited by the label predictor~\cite{sun2024eliminatinginformationleakagehard}. This architectural guarantee applies to raw image features bypassing the bottleneck, but the learned concept embeddings passed to the task head may carry residual information beyond the scalar concept value.

\paragraph{Proposition 2: Deterministic intervention propagation.}
Let each derived concept $c_j\in C_d$ be a deterministic function of its parents:
\begin{equation}
c_j = f_j(\mathrm{Pa}(c_j)).
\end{equation}
Then any intervention on a root concept $c_i\in C_r$, denoted $do(c_i=a)$, induces a unique
downstream concept state:
\begin{equation}
C_d^{\,do(c_i=a)}
=
F(C_r^{\,do(c_i=a)},M).
\end{equation}
Therefore, correcting a root concept automatically updates all downstream RANO quantities that depend on it.

\textbf{Interpretation.}
For example, if a clinician corrects $V_{t_1}$, then $\Delta V_{\%}$, $b_{\mathrm{PD}}^{\mathrm{vol}}$,
and $b_{\mathrm{PR}}^{\mathrm{vol}}$ are recomputed consistently. This is not guaranteed in a flat CBM, where all concepts are treated as independent predictors.
The proposition formalizes why our graph is suitable for test-time concept correction.

\subsection{TRACE Architecture}\label{sec:c2bm_architecture}

\begin{figure}[t]
  \centering
  \includegraphics[width=\linewidth]{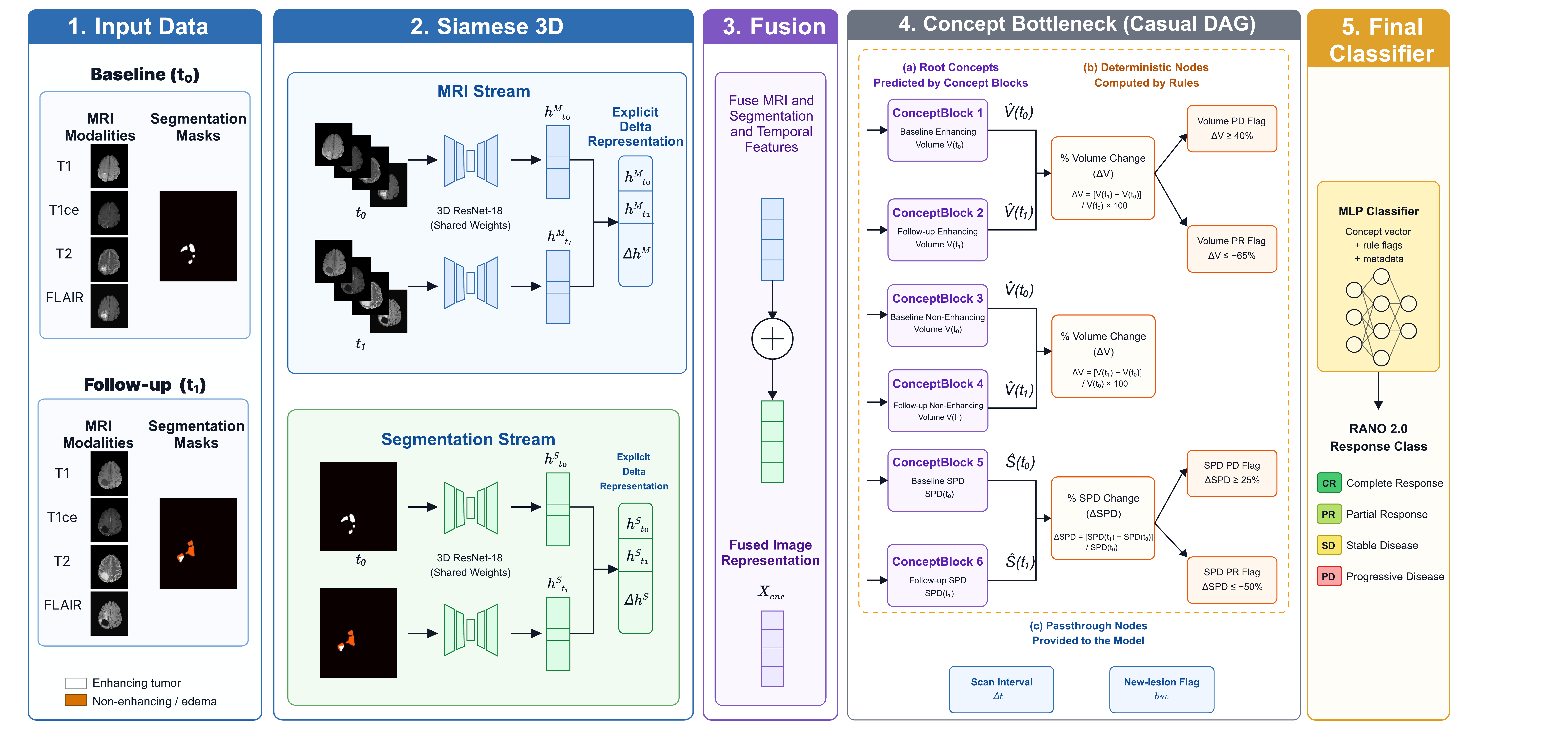}
  \caption{Overview of our proposed \textsc{Trace} CBM. (1) Baseline and follow-up MRI with segmentation masks are provided as input; (2) shared siamese 3D encoders extract MRI- and segmentation-based features and explicit delta representations; (3) these features are fused into a joint representation; (4) a causal concept bottleneck combines predicted root concepts, rule-based deterministic nodes, and passthrough clinical metadata; (5) a final MLP predicts the RANO 2.0 response category (CR, PR, SD, or PD).}
  \label{fig:c2bm_arch}
\end{figure}

Figure~\ref{fig:c2bm_arch} summarizes the proposed architecture for \textsc{Trace}. 
The model first encodes paired baseline and follow-up MRI volumes and segmentation masks into an image-only representation $\mathbf{X}_{\mathrm{enc}}$. This representation is used to predict root concepts. All downstream RANO concepts are then either computed deterministically or injected as passthrough metadata. Finally, a task head predicts the 4-class RANO response.

\subsubsection{Longitudinal Image Encoder}
\label{sec:longitudinal_encoder}

Image features are extracted using a siamese 3D CNN encoder. Two weight-sharing encoder instances process the baseline scan $t_0$, and follow-up scan $t_1$ separately. In our implementation, the encoder is a MedicalNet 3D ResNet-18~\cite{chen2019med3d}, pretrained on medical imaging tasks. The same framework can be instantiated with another 3D backbone.

We use two input streams; the first stream encodes the MRI volumes, where each scan is represented as a 4-channel tensor containing $T_1$, $T_1^{\mathrm{ce}}$, $T_2$, and FLAIR, resampled to $128\times128\times128$ voxels; and the second stream encodes the segmentation masks as a separate multi-channel input. Since the first convolutional layer of MedicalNet is originally designed for single-channel input, we adapt it to multi-channel input by repeating the pretrained weights across channels and dividing by the number of channels to preserve activation scale. For each input volume, global average pooling over the final feature map produces a 512-dimensional
feature vector:
\begin{equation}
\mathbf{h}
=
\mathrm{AdaptiveAvgPool3D}(f_{\theta}(x))
\in \mathbb{R}^{512}.
\end{equation}

For each stream, we represent the longitudinal pair using a temporal triplet:
\begin{equation}
\mathbf{T}(\mathbf{h}_{t_0},\mathbf{h}_{t_1})
=
\left[
\mathbf{h}_{t_0}
\;\|\;
\mathbf{h}_{t_1}
\;\|\;
\mathbf{h}_{t_1}-\mathbf{h}_{t_0}
\right]
\in \mathbb{R}^{3\times512}.
\label{eq:temporal_triplet}
\end{equation}

The MRI and segmentation triplets are concatenated and passed through a fusion layer $g_{\phi}$:
\begin{equation}
\mathbf{X}_{\mathrm{enc}}
=
g_{\phi}
\left(
[
\mathbf{T}^{\mathrm{MRI}}
\;\|\;
\mathbf{T}^{\mathrm{SEG}}
]
\right)
\in \mathbb{R}^{d},
\label{eq:fusion}
\end{equation}
where $g_{\phi}$ consists of a linear projection, layer normalization, and LeakyReLU activation.
The representation $\mathbf{X}_{\mathrm{enc}}$ is the only path through which image information enters
the concept bottleneck. Non-image metadata, such as $\Delta t$ and $b_{\mathrm{NL}}$, bypass the encoder and enter the graph only through passthrough nodes.

\subsubsection{Root Concept Prediction}
\label{sec:root_prediction}

The root concepts are the image-derived tumor measurements:
\begin{equation}
C_r =
\left\{
V_{t_0}, V_{t_1}, \mathrm{SPD}_{t_0}, \mathrm{SPD}_{t_1}
\right\}.
\end{equation}
Each root concept $c_k\in C_r$ is predicted from $\mathbf{X}_{\mathrm{enc}}$ by a concept-specific MLP, called a ConceptBlock. The ConceptBlocks share the same input representation but have independent parameters, allowing each concept head to specialize in its assigned clinical measurement. For a root concept $c_k$, the ConceptBlock computes
\begin{equation}
\mathbf{e}_k
=
\mathrm{MLP}_k(\mathbf{X}_{\mathrm{enc}})
\in \mathbb{R}^{d_c},
\end{equation}
followed by a scalar prediction
\begin{equation}
\hat{c}_k
=
\mathbf{w}_k^{\top}\mathbf{e}_k + b_k.
\end{equation}
The scalar prediction is trained against the z-scored ground-truth concept value. The embedding
$\mathbf{e}_k$ is retained as the representation of the root concept for downstream prediction.

\subsubsection{Graph Propagation}
\label{sec:graph_propagation}

The graph is represented by an adjacency matrix
\begin{equation}
\mathbf{G}\in\{0,1\}^{N\times N},
\end{equation}
where $N$ is the number of concept nodes plus the task node, and $G_{ij}=1$ denotes an edge
$c_i\rightarrow c_j$. The graph is topologically sorted so that root concepts are computed first, followed by deterministic derived concepts, and finally the task node. At level 0, root concepts are predicted from $\mathbf{X}_{\mathrm{enc}}$:
\begin{equation}
(\mathbf{e}_k,\hat{c}_k)
=
\mathrm{ConceptBlock}_k(\mathbf{X}_{\mathrm{enc}}),
\qquad
k\in C_r.
\end{equation}
These are the only concept nodes that receive image features directly. Intermediate nodes are not learned propagators; they are either deterministic RANO nodes or passthrough nodes. This design reflects the structure of the clinical decision rule, i.e., once the root measurements and metadata are known, percentage changes and RANO threshold indicators are fully specified by fixed computations. Additional implementation details for the ConceptBlock formulation are provided in Appendix~\ref{app:concept_block_details}.

\subsubsection{Deterministic RANO Nodes}
\label{sec:deterministic_nodes}

Deterministic nodes implement RANO quantities that are completely determined by their parents.
These nodes have neither ConceptBlocks nor trainable parameters. Since root concepts are predicted in
z-scored space, we first convert them back to the clinical scale before applying RANO computations.
Let $\mu_k$ and $\sigma_k$ denote the training-fold mean and standard deviation of concept $c_k$.
The raw-scale prediction is
\begin{equation}
\tilde{c}_k
=
\mathrm{expm1}(\hat{c}_k\sigma_k+\mu_k),
\label{eq:raw_scale}
\end{equation}
where the $\mathrm{expm1}$ transformation is used when the concept labels were log-transformed during
preprocessing. For example, the volume-based derived concepts are computed as
\begin{align}
\widehat{\Delta V}_{\%}
&=
100\cdot
\mathrm{clamp}
\left(
\frac{\tilde{V}_{t_1}-\tilde{V}_{t_0}}
{\tilde{V}_{t_0}+\varepsilon},
-1,5
\right), \\
\hat{b}_{\mathrm{PD}}^{\mathrm{vol}}
&=
\mathbf{1}
\left[
\widehat{\Delta V}_{\%}\geq 40
\;\wedge\;
\tilde{V}_{t_0}\geq 0.5
\right], \\
\hat{b}_{\mathrm{PR}}^{\mathrm{vol}}
&=
\mathbf{1}
\left[
\widehat{\Delta V}_{\%}\leq -65
\;\wedge\;
\tilde{V}_{t_0}\geq 0.5
\right].
\end{align}
The SPD-derived concepts are computed analogously using the thresholds in Table~\ref{tab:rano_criteria}.
Deterministic nodes are excluded from the concept supervision loss because their values are fixed
functions of their parents.
Hard threshold indicators near the RANO boundaries could be replaced by a temperature-controlled sigmoid relaxation $\sigma((u-\tau)/T)$ to provide smoother gradients. In practice, the hard-threshold formulation was stable, so we use it as the default. Additional implementation details, including normalization, clipping, and the SPD-derived flags, are provided in Appendix~\ref{app:deterministic_nodes}.

\subsubsection{Passthrough Metadata Nodes}
\label{sec:passthrough_nodes}

Two concepts are treated as passthrough nodes: the scan interval $\Delta t$ and the new-lesion flag
$b_{\mathrm{NL}}$. These quantities are not predicted from image features; instead, their values are injected directly into the graph:
\begin{equation}
\hat{c}_k = c_k^{*},
\qquad
k\in\{\Delta t,b_{\mathrm{NL}}\}.
\end{equation}

This choice is clinically motivated because the scan interval is administrative metadata, while the new-lesion flag is computed from lesion topology across the two time points. Including these concepts as explicit passthrough nodes prevents the image encoder from being forced to represent non-image information, while still allowing the RANO graph to use clinically relevant context. Further details on the construction of the new-lesion flag and the handling of passthrough nodes are provided in Appendix~\ref{app:passthrough_nodes}.

\subsubsection{Task Prediction}
\label{sec:task_prediction}

The task node $Y$ predicts the 4-class RANO response. It receives the outputs of its parent concepts, including root concept embeddings and scalar deterministic or passthrough values. 
We denote these inputs by $\mathbf{z}_{p_1},\ldots,\mathbf{z}_{p_m}$. 
The prediction is
\begin{equation}
\hat{\mathbf{y}}
=
\mathrm{softmax}
\left(
\mathrm{MLP}_{Y}
\left(
[
\mathbf{z}_{p_1}
\;\|\;
\cdots
\;\|\;
\mathbf{z}_{p_m}
]
\right)
\right),
\label{eq:task_prediction}
\end{equation}
where $\hat{\mathbf{y}}\in\mathbb{R}^{4}$ is a probability distribution over
$\{\mathrm{CR},\mathrm{PR},\mathrm{SD},\mathrm{PD}\}$.

Although the task head is learned, it is constrained to operate only over the structured bottleneck.
This makes its input inspectable and allows clinicians to apply corrections at the concept level.

\subsection{Training Objective}
\label{sec:training_objective}

The model is trained with a combination of task supervision, deep-supervision on the dominant PD decision, root-concept supervision, intervention-consistency regularization, and a small $L_1$ penalty on the first layer of every learnable concept propagator:
\begin{equation}
\mathcal{L}_{\mathrm{total}}
=
\mathcal{L}_{\mathrm{task}}
+
\alpha_{\mathrm{aux}}\mathcal{L}_{\mathrm{aux}}
+
\lambda_c \mathcal{L}_{\mathrm{concept}}
+
\beta_{\mathrm{cons}}\mathcal{L}_{\mathrm{IC}}
+
\gamma\,\big\lVert \theta_{\mathrm{prop}}^{(1)} \big\rVert_1.
\label{eq:total_loss}
\end{equation}
In our final configuration $\lambda_c=0.8$, $\alpha_{\mathrm{aux}}=0.1$, $\beta_{\mathrm{cons}}=0.5$,
and $\gamma=10^{-4}$. The $L_1$ term targets only the input layer of each propagator so that the
bottleneck cannot route image information into the task head through a few high-magnitude weights; it acts as a shortcut suppressor rather than a generic regularizer.

\paragraph{Task loss.}
The primary task loss is a soft macro F1 objective combined with focal weighting \cite{Lin_2017_ICCV}, which directly optimizes the imbalanced evaluation metric:
\begin{equation}
\mathcal{L}_{\mathrm{task}}
=
1
-
\frac{1}{4}
\sum_{j=1}^{4}
\frac{2\sum_{i} (1-\hat{y}_{ij})^{\gamma_f}\, y_{ij}\,\hat{y}_{ij}}
     {\sum_{i}\big[(1-\hat{y}_{ij})^{\gamma_f}\,y_{ij} + \hat{y}_{ij}\big] + \varepsilon},
\label{eq:task_loss}
\end{equation}
where $\gamma_f$ is the focal exponent applied to confident predictions and $\varepsilon$ stabilizes the ratio for classes with few positives in a batch. The class-weighted cross-entropy $-\sum_j w_j y_j \log \hat{y}_j$ with inverse-frequency weights $w_j$ is retained only as an ablation baseline.

\paragraph{Auxiliary PD head.}
The four RANO outcomes collapse into a clinically dominant PD-vs-rest decision. We add a
deep-supervision auxiliary head that reads the same task logits and predicts the binary PD probability
$\hat{y}_{\mathrm{PD}}^{\mathrm{bin}}=\hat{y}_{\mathrm{PD}}$, supervised by class-weighted binary
cross-entropy:
\begin{equation}
\mathcal{L}_{\mathrm{aux}}
=
-\big[
w_{+}\,y_{\mathrm{PD}}^{\mathrm{bin}}\log \hat{y}_{\mathrm{PD}}^{\mathrm{bin}}
+
w_{-}\,(1-y_{\mathrm{PD}}^{\mathrm{bin}})\log(1-\hat{y}_{\mathrm{PD}}^{\mathrm{bin}})
\big],
\end{equation}
with $w_{+},w_{-}$ inverse-frequency weights. This term provides a stable supervision signal for the most frequent clinical decision and indirectly stabilizes the rest of the bottleneck.

\paragraph{Concept loss.}
Concept supervision is applied only to learnable root concepts:
\begin{equation}
\mathcal{L}_{\mathrm{concept}}
=
\frac{1}{|C_r|}
\sum_{c_k\in C_r}
\mathrm{MSE}(\hat{c}_k,c_k^{*}).
\label{eq:concept_loss}
\end{equation}
Derived and passthrough nodes are excluded because their values are fixed functions of their parents or are injected directly. Concept supervision prevents the bottleneck from collapsing into arbitrary latent codes and keeps each ConceptBlock aligned with its assigned clinical measurement.

\paragraph{Encoder freezing.}
The 3D encoder is fine-tuned jointly with the concept and task heads during the first phase of training, and frozen at epoch~50 for the remainder of training. This stabilizes the concept-to-task mapping by preventing late-stage drift of the shared image representation.

\subsection{Intervention-Consistency Training}
\label{sec:intervention_consistency}

A known failure mode of CBMs is that the task head adapts to systematic errors in the predicted
concept distribution; when those concepts are replaced by correct values at test time, the input becomes distribution-shifted, and the final prediction degrades \cite{NEURIPS2022_944ecf65}. To prevent this train-test mismatch, we train the task head jointly under predicted and intervened concept states, and explicitly penalize disagreement between the two forward passes. Ablation results (In the upcoming Section~\ref{sec:ablation_results}) show that it is the largest single contributor to performance. During training, each root concept prediction is randomly replaced by its ground-truth value with probability $p_{\mathrm{int}}$:
\begin{equation}
\tilde{c}_k
=
m_k c_k^{*}
+
(1-m_k)\hat{c}_k,
\qquad
m_k\sim \mathrm{Bernoulli}(p_{\mathrm{int}}).
\label{eq:mixed_concepts}
\end{equation}
The deterministic downstream RANO concepts are then recomputed from the mixed root state,
$\tilde{C}_d = F(\tilde{C}_r,M)$, and the intervened prediction is
$\tilde{\mathbf{y}} = g_Y(\tilde{C}_r,\tilde{C}_d,M)$. We penalize disagreement between the
fully predicted and the intervened predictions with the Kullback–Leibler (KL) divergence term
\begin{equation}
\mathcal{L}_{\mathrm{IC}}
=
D_{\mathrm{KL}}
\left(
\hat{\mathbf{y}}
\;\|\;
\tilde{\mathbf{y}}
\right),
\label{eq:ic_loss}
\end{equation}
where $\hat{\mathbf{y}} = g_Y(\hat{C}_r,\hat{C}_d,M)$. We anneal the intervention probability linearly
over the first 15 epochs,
\begin{equation}
p_{\mathrm{int}}(e)
=
\min\!\left(p_{\max},\;
p_{\min} + (p_{\max}-p_{\min})\cdot\tfrac{e}{E_{\mathrm{anneal}}}\right),
\qquad
p_{\min}=0.3,\;p_{\max}=0.8,\;E_{\mathrm{anneal}}=15,
\label{eq:pint_schedule}
\end{equation}
so that the task head first learns from largely predicted concepts and then increasingly relies on
corrected ones. This schedule, together with the KL term, makes the model robust to clinically
plausible concept corrections at test time, which is the practical setting in which a clinician can audit or override an intermediate RANO measurement.
We also considered monotonicity constraints on the task head, but treat them as an optional extension rather than part of the final model; details are provided in Appendix~\ref{app:monotonicity_constraints}.

\subsection{Adapting to Longitudinal RANO Assessment}
\label{sec:c2bm_adaptation}

Our model builds on the causal CBM from \citet{defelice2026causallyreliableconceptbottleneck}, but adapts the framework to longitudinal GBM response assessment. 
We extended the original CBM from single-image prediction to paired baseline and follow-up multimodal MRI. A siamese 3D encoder processes both time points, together with their segmentation masks, and represents the longitudinal change through feature-level differences. This allows the model to predict clinically meaningful root concepts from volumetric temporal inputs rather than from isolated 2D images.

The graph is also adapted to the RANO setting. Rather than learning the concept structure from data, we manually specify the DAG from the RANO decision process described in Section~\ref{sec:rano_criteria}. Image information enters the bottleneck only through the root concepts, while downstream RANO concepts, such as percentage changes and threshold flags, are computed deterministically from their parents. This prevents derived concepts from carrying hidden image embeddings to the task head and keeps the reasoning pathway auditable. We also use intervention-consistency training, with corrected concept values shown to the task head during training, so that concept-level corrections can propagate more reliably to the final response prediction. Overall, the method converts longitudinal RANO classification into a structured concept reasoning problem designed for small and imbalanced neuro-oncology datasets.

\section{Experiments}\label{sec:experiments}

\subsection{Dataset}\label{sec:dataset}

For our experiments, we selected the LUMIERE dataset~\cite{suter2022lumiere} because it provides scan-pair-level RANO response labels assigned through formal response assessment, covering the four response categories CR, PR, SD, and PD. 
This is important for our setting because \textsc{Trace} is designed around the RANO decision process. The model uses image-derived measurements, deterministic threshold flags, and passthrough clinical metadata, so the training labels must match the same response assessment framework. 
We also considered using other datasets, for example~\cite{gagnon2025ucsd,fields2024ucsf}, but did not use them for training because they do not provide equivalent RANO annotations at the scan-pair level. Reconstructing RANO labels from segmentation-derived volume change introduced label ambiguity rather than a like-for-like supervision signal. Further details are provided in Appendix~\ref{app:external_datasets}.

\begin{figure} 
\centering
\includegraphics[width=0.85\linewidth]{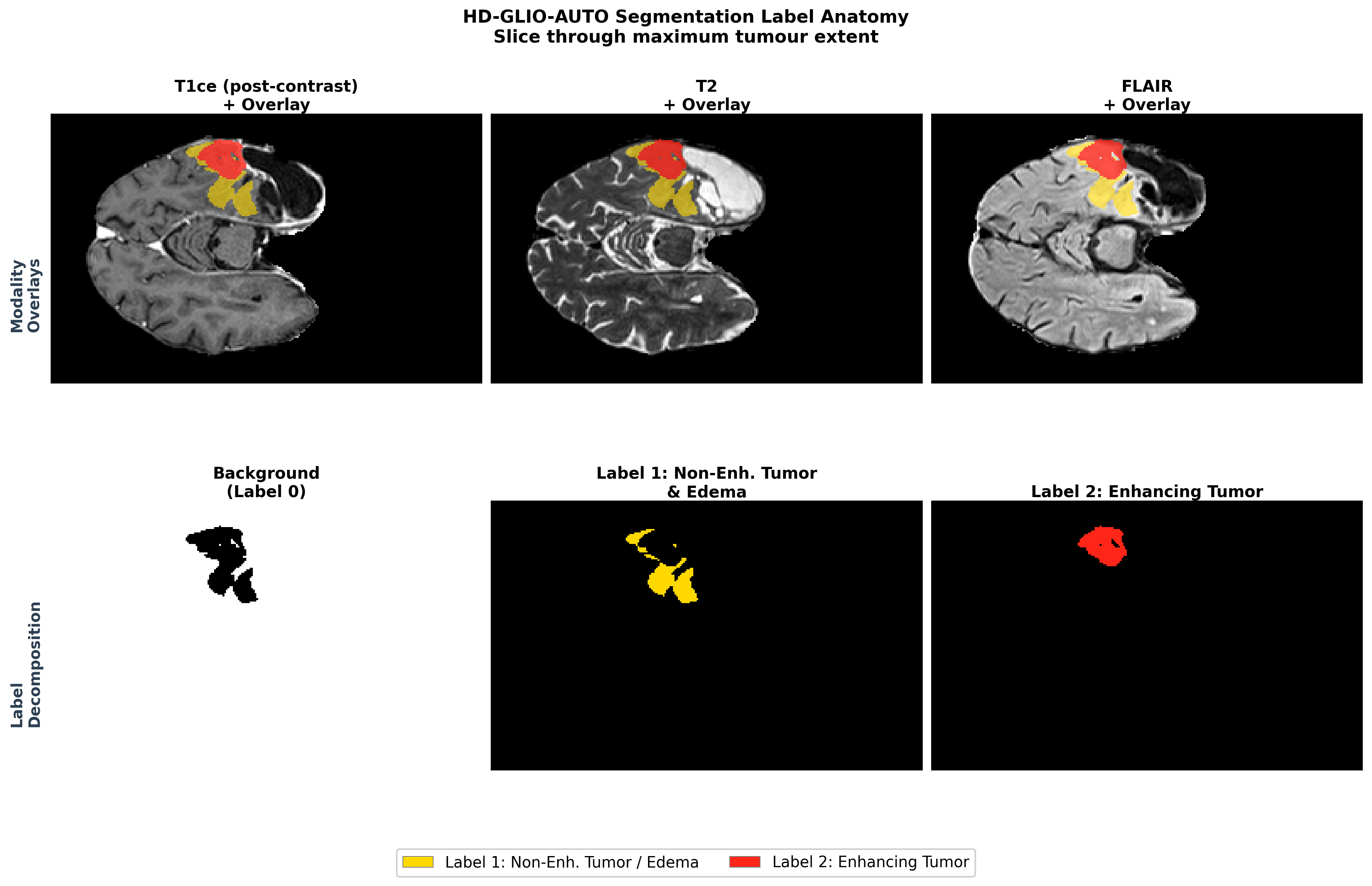}
\caption{Representative segmentation case from LUMIERE. Top row: three MRI modalities with the HD-GLIO-AUTO overlay. Bottom row: non-enhancing (yellow) and enhancing (red) tumor labels in isolation. 
These regions are used to compute volumetric concepts for \textsc{Trace}.
}
\label{fig:seg_anatomy}
\end{figure}

LUMIERE contains data from 91 patients. 
We ended up with 599 scan sessions after filtering for complete modality sets and valid longitudinal follow-ups. Baseline and follow-up pairs are constructed from patients with at least two valid consecutive time points, resulting in 361 longitudinal pairs, with a mean of 4.8 visits per patient and a median follow-up of 53 weeks. 
Each session contains four spatially aligned, skull-stripped 3D MRI sequences (T1, T1ce, T2, FLAIR) resampled to $1{\times}1{\times}1$\,mm and an automatically generated tumor segmentation from HD-GLIO-AUTO~\cite{kickingereder2019automated} with three labels: background, non-enhancing tumor, and contrast-enhancing tumor. 
The segments are used to compute the volumetric concepts described in Sections~\ref{sec:rano_criteria} and~\ref{sec:concept_structure}. 
Figure~\ref{fig:seg_anatomy} shows an example segmentation case.

\subsubsection{Preprocessing}\label{sec:DataPreprocessing}

We applied a four-step preprocessing pipeline to improve spatial and intensity consistency across patients and longitudinal scans. First, all MRI sequences were registered to the MNI152 1\,mm template to support voxel-wise correspondence across subjects. Second, intensity inhomogeneity was corrected using N4 bias field correction~\cite{tustison2010n4itk}. Third, histogram matching was applied to reduce inter-scan intensity variation. Finally, each volume was z-score normalized. 
Segmentation masks were transformed to the same template space using the MRI-to-template transformation and resampled with nearest-neighbor interpolation to preserve discrete labels. The pipeline follows the approach in~\cite{tikhonov2025predicting}, with segmentation handling adapted to the concept-based framework. Additional details are provided in Appendix~\ref{app:preprocessing_details}.

\subsubsection{Label Distribution}\label{sec:rano_distribution}

\begin{figure}[ht]
\centering
\includegraphics[width=0.50\linewidth]{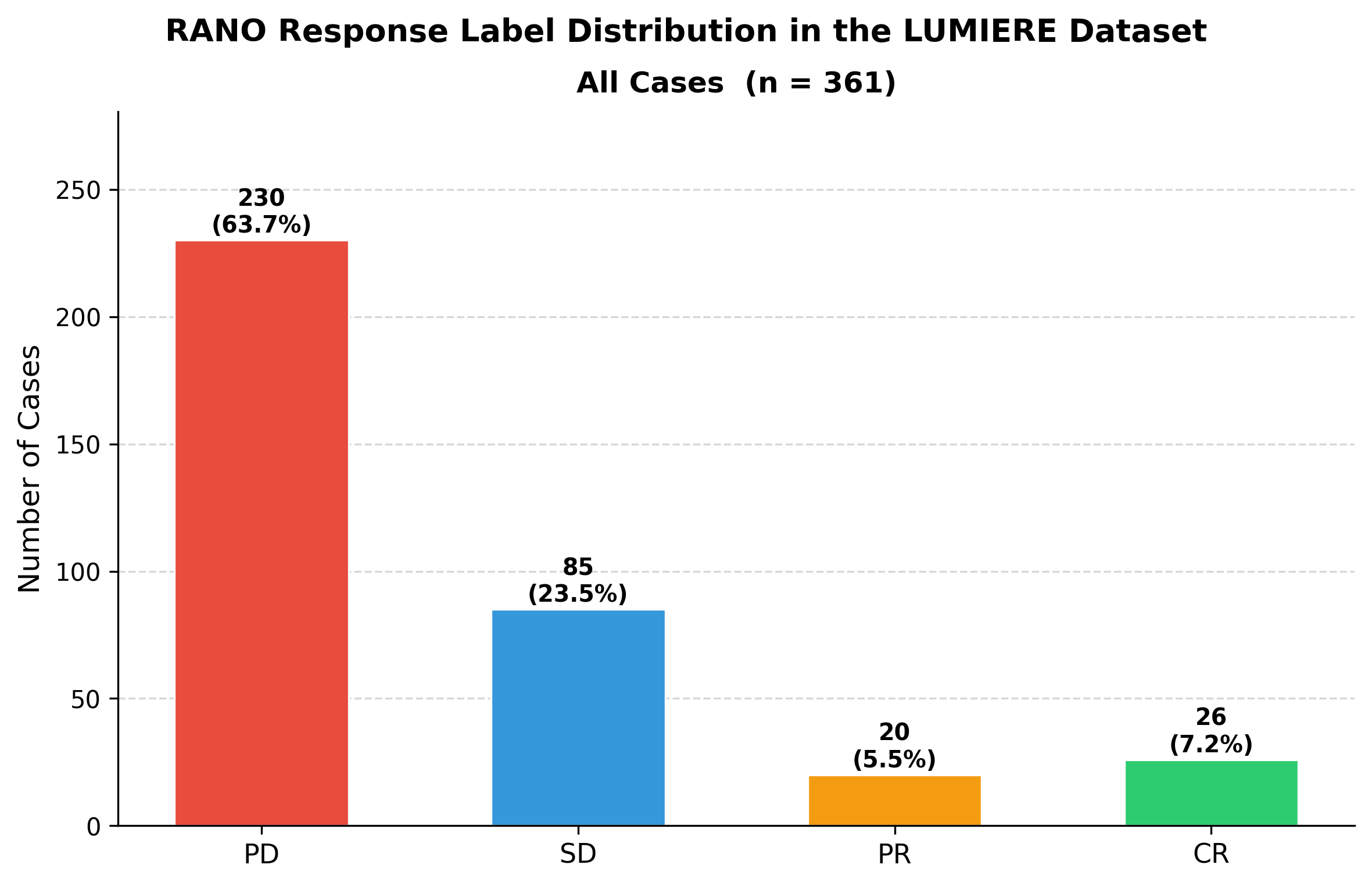}
\caption{RANO label distribution across all 361 baseline--follow-up pairs in LUMIERE. PD accounts for 63.7\% of samples, leading to a strongly imbalanced 4-class classification task.}
\label{fig:rano_dist}
\end{figure}

The RANO label distribution is strongly imbalanced, as shown in Figure~\ref{fig:rano_dist}, with PD accounting for 63.7\% of the 361 pairs. 
This imbalance affects both training and evaluation, increasing the uncertainty of per-class estimates and making plain accuracy less informative. Therefore, we use macro F1 as the primary metric and address class imbalance during training, as described in Section~\ref{sec:experimental_setup}. The small number of longitudinal samples also motivates the use of a pretrained 3D encoder and concept-level supervision rather than training a large 3D classifier from scratch.

\subsection{Baseline Models}\label{sec:BaselineModels}

We compare \textsc{Trace} against two baseline groups. 

\paragraph{Black-Box Deep Learning Baselines.} 
First, against three published black-box deep learning baselines evaluated on LUMIERE~\cite{matoso2025deeplearningapproachclassifying,tikhonov2025predicting,amato2025integrating}.
For \citet{matoso2025deeplearningapproachclassifying}, we use the replicated values reported by \citet{amato2025integrating}, as they follow the same data splits as our experiments. 
For \citet{tikhonov2025predicting}, we report the published values, which were consistent with our reproduction.
These baselines provide useful performance references, but they do not encode the RANO decision process or support the same concept-level inspection and correction as \textsc{Trace}.

\paragraph{Interpretable Baselines.}
Second, against interpretable baselines, including our implementation of DCBM~\cite{prasse2025dcbmdataefficientvisualconcept} and two rule-based RANO oracle references. DCBM tests whether automatically discovered concepts from frozen 3D medical embeddings can support longitudinal RANO response classification without explicit concept supervision; we adapt DCBM to paired MRI scans by extracting patch-level embeddings at baseline and follow-up, learning sparse concept activations, and classifying longitudinal changes in those activations; implementation details are provided in Appendix~\ref{app:dcbm_details}. The RANO oracle references apply deterministic threshold rules to either ground-truth concepts (GT concept) or predicted concepts (predicted concepts), isolating the behaviour of threshold-only decision logic and its sensitivity to concept-estimation noise.

\subsection{Experimental Setup}\label{sec:experimental_setup}

We evaluate \textsc{Trace} on both 4-class RANO response classification and binary progression-versus-non-progression classification, where PD is compared against all non-PD responses.

\paragraph{Cross-validation.} 
We used 5-fold Stratified Group K-Fold cross-validation. Stratification is applied at the scan-pair level to preserve the RANO class distribution across folds, while grouping ensures that scans from the same patient do not appear in both the training and validation splits.

\paragraph{Backbone and inputs.} 
The \textsc{Trace} image encoder is a 3D MedicalNet ResNet-18~\cite{chen2019med3d}, with the four MRI modalities and three segmentation channels provided at $128{\times}128{\times}128$ resolution, in bf16 mixed precision. 
The encoder is fine-tuned jointly with the concept and task heads during the first phase of training and frozen at epoch 50 to stabilize the concept-to-task mapping.

\paragraph{Loss and intervention training.} 
We train with a soft macro F1 loss combined with focal weighting, so that optimization is better aligned with the primary metric under class imbalance.
We also add a deep-supervision auxiliary binary head on the PD probability of the 4-class output, with weight $\alpha_{\mathrm{aux}}=0.1$, to provide a stable PD-vs-rest supervision signal. Intervention-aware training~\cite{defelice2026causallyreliableconceptbottleneck} is used with intervention probability $p_{\mathrm{int}}$ annealed from $0.3$ to $0.8$ over the first 15 epochs. An intervention-consistency KL term between the predicted-concept and intervened-concept forward passes is added with weight $\beta_{\mathrm{cons}}=0.5$. We apply an $L_1$ penalty of $10^{-4}$ to the first layer of the task head to discourage reliance on a small number of high-magnitude bottleneck inputs. 

\paragraph{Optimization.} 
All variants are trained with AdamW~\cite{loshchilov2019decoupledweightdecayregularization}, learning rate $10^{-3}$, $L_2$ weight decay $0.01$, batch size $16$, and a maximum of $100$ epochs. A \texttt{ReduceLROnPlateau} scheduler monitors validation macro F1, and early stopping is applied after $20$ epochs without improvement. The concept loss weight is $\lambda_c=0.8$, and the checkpoint with the best validation macro F1 is retained. Full hyperparameters are listed in Appendix~\ref{app:training_hyperparams}.

\paragraph{Metrics.} 
We used macro F1 over the four RANO classes as the primary metric. We also report weighted accuracy, per-class F1, and binary PD-vs-rest macro F1 to assess performance beyond the aggregate 4-class score. For interpretability, we report CaCE-TV \cite{goyal2019explaining}, Concept-Alignment Score, and an Ordered Intervention Score; these metrics measure concept influence, concept prediction quality, and the effect of ordered concept interventions, respectively.

\subsection{Ablation Studies}\label{sec:ablation_studies}

We used ablation studies to test which parts of \textsc{Trace} contribute to performance. 
The ablations focus on two aspects of the model, i.e., the training objective and the expert RANO graph. All variants are evaluated on fold~4, which is the strongest individual fold and provides a consistent reference for comparing effect direction. Each ablation starts from the full configuration and disables one component while keeping the remaining hyperparameters fixed. Additional ordered-versus-random intervention policy ablation results are provided in Appendix~\ref{app:intervention_policy_ablation}.

\begin{itemize}\setlength\itemsep{0.2em}
  \item \textbf{A0 (full model):} uses the expert RANO DAG, soft macro F1 with focal weighting, the auxiliary PD head, intervention-consistency KL, the deterministic derivation chain, and $L_1$ regularization on the first layer of the task head.

  \item \textbf{A1 ($-$consistency KL):} sets $\beta_{\mathrm{cons}}=0$. This tests whether aligning the predicted-concept and intervened-concept forward passes is needed to maintain a stable bottleneck under concept correction.

  \item \textbf{A3 ($-$auxiliary PD head):} sets $\alpha_{\mathrm{aux}}=0$. This tests whether the PD-vs-rest deep-supervision signal contributes beyond the 4-class task loss.

  \item \textbf{A4 (cross-entropy task loss):} replaces the soft macro F1 with focal weighting by plain cross-entropy. This tests whether the macro F1-aligned task loss improves performance under the strong class imbalance in LUMIERE.

  \item \textbf{B1 (bipartite graph):} replaces the expert RANO DAG with a flat bipartite CBM. In this setting, the deterministic derivation chain and passthrough nodes are disabled because the bipartite graph does not encode the parent-child dependencies required for closed-form RANO computation. This tests whether the clinical dependency structure adds value beyond exposing the same concepts to the classifier.

  \item \textbf{B2 ($-$deterministic chain):} keeps the expert DAG but disables the deterministic derivation chain, forcing derived concepts to be predicted by learnable graph modules. This isolates the value of deterministic RANO computation from the value of the graph structure itself.
\end{itemize}

\section{Results and Discussion}\label{sec:results_discussion}

We evaluate \textsc{Trace} on 4-class RANO treatment-response classification using 5-fold patient-wise cross-validation on the LUMIERE dataset~\cite{suter2022lumiere}. Across folds, \textsc{Trace} achieves a 4-class macro F1 of $0.4769 \pm 0.1229$ and a weighted accuracy of $0.5447 \pm 0.1318$. On the clinically important PD-vs-rest binary decision, macro F1 reaches $0.7085 \pm 0.0935$, and weighted accuracy reaches $0.7329 \pm 0.1104$. 
The main source of variation is the small number of CR and PR cases in each fold; the detailed fold-level results are provided in Appendix~\ref{app:fold_metrics}. 
Figure~\ref{fig:perclass_f1} shows that PD achieves the highest and most consistent F1 across folds, whereas PR has the largest fold-to-fold variation, reflecting the small number of PR cases in several validation folds. The pooled confusion matrix in Figure~\ref{fig:confusion_pooled} shows that most errors occur between clinically adjacent response categories, especially SD versus PD and CR/PR versus SD. This is consistent with the RANO task itself, where class boundaries depend on continuous tumor-burden measurements and threshold-based changes that can be sensitive to segmentation and measurement noise. 
The full per-class table is provided in Appendix~\ref{app:per_class_metrics}.

\begin{figure}[t]
    \centering
    \subfloat[Per-class F1 across folds.\label{fig:perclass_f1}]{
        \includegraphics[width=0.47\textwidth]{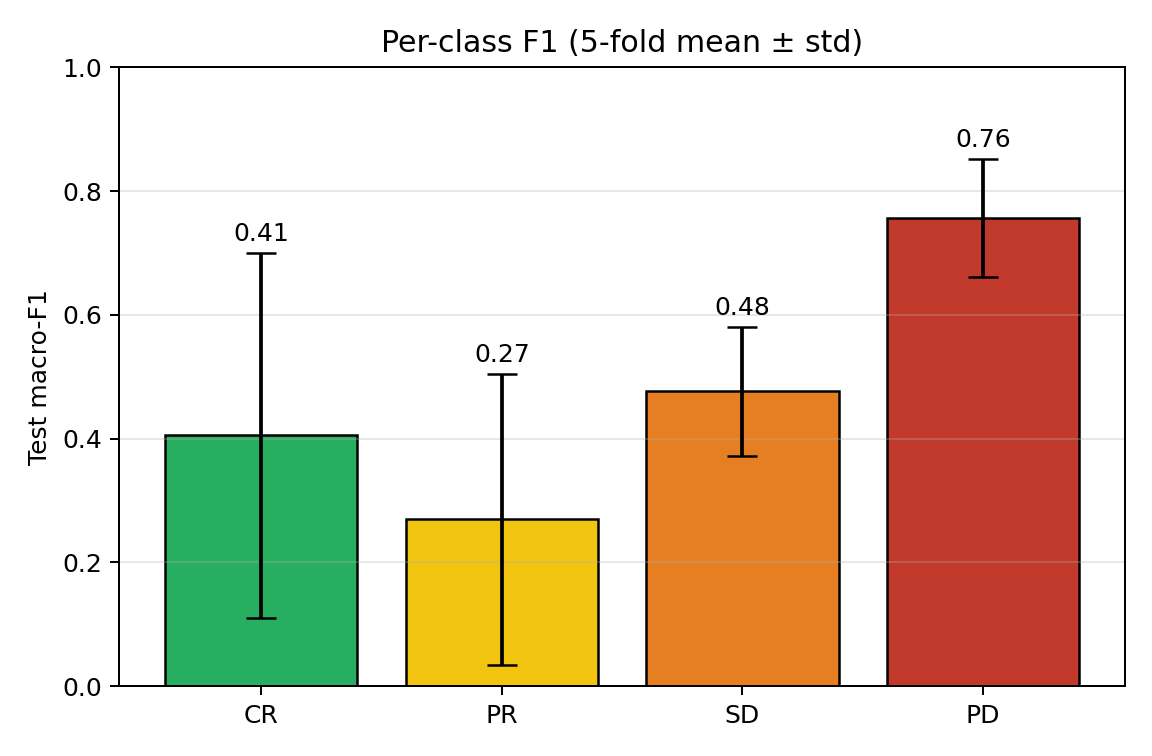}
    }
    \hfill
    \subfloat[Pooled five-fold confusion matrix.\label{fig:confusion_pooled}]{
        \includegraphics[width=0.47\textwidth]{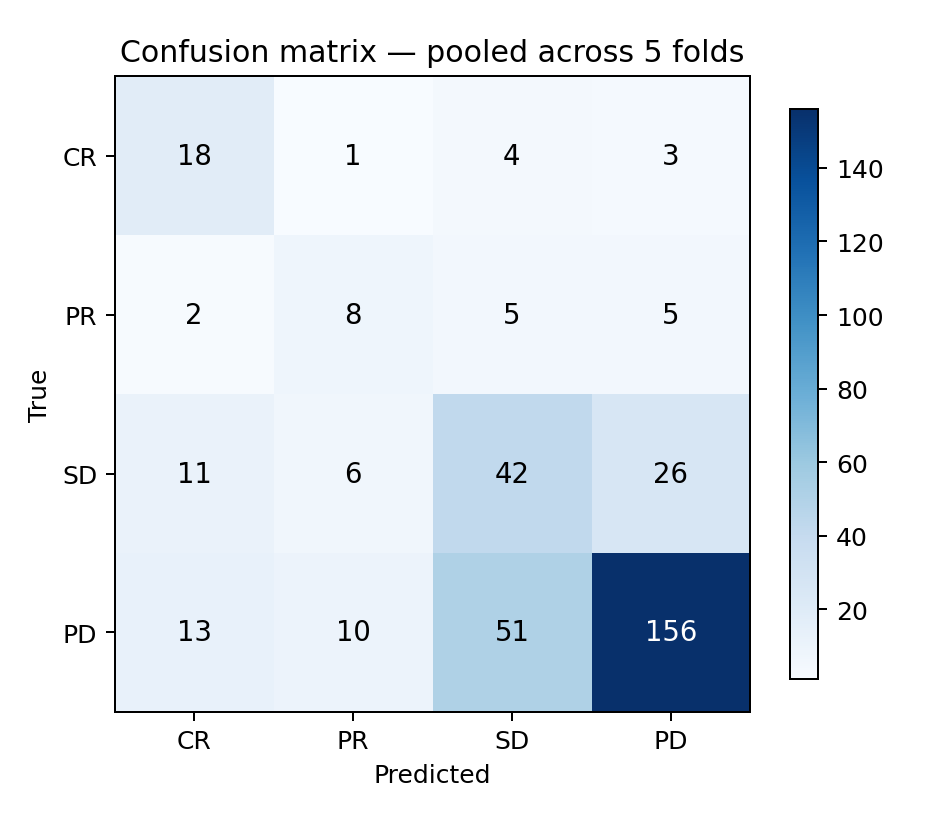}
    }
    \caption{Per-class performance and pooled errors on LUMIERE.}
    \label{fig:perclass_and_confusion}
\end{figure}

\subsection{Baseline Comparison.}
Table~\ref{tab:main_comparison} compares \textsc{Trace} with published black-box baselines~\cite{matoso2025deeplearningapproachclassifying,amato2025integrating,tikhonov2025predicting} and the concept-based DCBM baseline~\cite{prasse2025dcbmdataefficientvisualconcept}. We also report a \emph{rule-based RANO oracle}, which applies the textbook RANO decision tree directly to the ground-truth concept vector for each validation pair. The oracle assigns PD when a new lesion is present, $\Delta V_{\mathrm{enh}}\geq 40\%$, or $\Delta\mathrm{SPD}\geq25\%$, with analogous threshold rules for PR and CR; this provides a threshold-only reference with perfect access to the concepts that \textsc{Trace} must estimate from MRI, but without a learned task head.

Despite using ground-truth concepts, the Rule-based RANO oracle (GT concept) reaches only $0.346\pm0.038$ 4-class macro F1 and $0.681\pm0.058$ binary macro F1, which is $13$ percentage points below \textsc{Trace} on the 4-class task. It also recovers none of the $26$ ground-truth CR cases and misses $41\%$ of PD cases across the five folds. This suggests that the LUMIERE labels are not fully determined by segmentation-derived thresholds alone, likely because expert RANO assessment can also reflect non-measurable disease, steroid status, clinical deterioration, and radiologist judgment. 
Using the same threshold rules, the Rule-based RANO oracle (predicted concepts) further reduces macro F1 to $0.238\pm0.041$, showing that hard-threshold decisions are sensitive to concept estimation noise.

\textsc{Trace} also improves over DCBM, increasing macro F1 from $0.254$ to $0.4769$ and weighted accuracy from $0.313$ to $0.5447$. 
The strongest black-box baselines retain a macro F1 advantage~\cite{amato2025integrating,tikhonov2025predicting}, but they rely on direct label-discriminative image or radiomic features and do not constrain prediction through an auditable RANO-aligned concept pathway. The DenseNet-264 result~\cite{matoso2025deeplearningapproachclassifying} further shows that end-to-end classifier capacity alone is not enough to resolve the class imbalance in this dataset. Overall, \textsc{Trace} offers a different trade-off as it improves over concept-based alternatives while preserving concept-level inspection and correction.

\begin{table}[ht]
    \centering
    \caption{RANO classification on LUMIERE reported as mean $\pm$ std. Values are rounded to three decimal places. 
    The results reported by \citet{matoso2025deeplearningapproachclassifying} are based on the replicated values from \citet{amato2025integrating}. 
    The accuracy type varies across methods due to differences in how published results are reported; therefore, direct numerical comparisons should be treated with caution.}
    \label{tab:main_comparison}
    \resizebox{\textwidth}{!}{%
    \begin{tabular}{lcclc}
        \toprule
        \textbf{Model} & \textbf{Macro F1} & \multicolumn{2}{c}{\textbf{Accuracy Score}} & \textbf{Interpretable} \\
        \midrule
        \citet{matoso2025deeplearningapproachclassifying} (+ Radiomic Features)
        & 0.505 $\pm$ 0.098
        & 0.470 $\pm$ 0.091
        & Balanced acc.
        & $\times$ \\
        
        \citet{amato2025integrating}
        & 0.616 $\pm$ 0.048
        & 0.322 $\pm$ 0.055
        & Balanced acc.
        & $\times$ \\
        
        \citet{tikhonov2025predicting}
        & 0.500 $\pm$ 0.080
        & 0.720 $\pm$ 0.050
        & Accuracy
        & $\times$ \\
        
        DCBM Baseline
        & 0.254 $\pm$ 0.040
        & 0.313 $\pm$ 0.070
        & Balanced acc.
        & $\checkmark$ \\
        
        Rule-based RANO oracle (GT concepts)
        & 0.346 $\pm$ 0.038
        & 0.601 $\pm$ 0.037
        & Accuracy
        & $\checkmark$ \\

        Rule-based RANO oracle (predicted concepts)
        & 0.238 $\pm$ 0.041
        & 0.645 $\pm$ 0.044
        & Accuracy
        & $\checkmark$ \\
        \midrule
        \textbf{\textsc{Trace} (Our Approach)}
        & \textbf{0.477 $\pm$ 0.123}
        & \textbf{0.545 $\pm$ 0.132}
        & \textbf{Weighted acc.}
        & \textbf{$\checkmark$} \\
        \bottomrule
    \end{tabular}}
\end{table}

\subsection{Ablation Results}\label{sec:ablation_results}

We use the fold~4 ablations in Table~\ref{tab:ablation} to examine which parts of \textsc{Trace} contribute most to performance. Each variant starts from the full model A0 and disables one component while keeping the remaining hyperparameters fixed, as explained in Section~\ref{sec:ablation_studies}. The ablations focus on two parts of the design, which are the training objective and the expert RANO graph.

\begin{table}[ht]
    \centering
    \caption{Component ablation on fold~4. A0 is the full model; each row disables one component. Negative $\Delta$ indicates the drop relative to A0.}
    \label{tab:ablation}
    \setlength{\tabcolsep}{5pt}
    \small
    \begin{tabular}{llccc}
        \toprule
        \textbf{ID} & \textbf{Variant} & \textbf{4-class F1} & \textbf{Binary F1} & \textbf{$\Delta$ F1 vs A0} \\
        \midrule
        A0 & Full model & 0.6963 & 0.8572 & -- \\
        A1 & $-$intervention-consistency KL & 0.3538 & 0.6000 & $-0.342$ \\
        A3 & $-$auxiliary PD head ($\alpha_{\mathrm{aux}}=0$) & 0.6078 & 0.7565 & $-0.089$ \\
        A4 & Cross-entropy task loss & 0.6635 & 0.7159 & $-0.033$ \\
        B1 & Bipartite graph & 0.3549 & 0.5921 & $-0.342$ \\
        B2 & Expert DAG without deterministic derivation & 0.3214 & 0.5774 & $-0.375$ \\
        \bottomrule
    \end{tabular}
\end{table}

\noindent \textbf{Core bottleneck components.}
The largest drops occur when removing intervention-consistency training or disrupting the expert RANO graph. Removing the intervention-consistency KL term (A1) reduces 4-class macro F1 from $0.6963$ to $0.3538$ and binary macro F1 from $0.8572$ to $0.60$; this supports the role of the consistency term in reducing the mismatch between predicted-concept inputs and intervened-concept inputs, which is important for stable test-time concept correction. Replacing the expert RANO DAG with a flat bipartite graph (B1) produces a similar reduction, while keeping the expert DAG but disabling deterministic RANO derivation (B2) gives the largest drop, reducing 4-class macro F1 to $0.3214$.

The graph ablations clarify the role of the structured bottleneck. B1 shows that exposing the concepts to the classifier is not sufficient when the clinical dependency structure is removed. B2 shows that the expert graph is also not sufficient if derived RANO quantities are learned rather than computed deterministically from their parent concepts. In other words, the main benefit comes from combining the graph structure with deterministic RANO computation, because root measurements are estimated from MRI, while percentage changes and threshold flags are recomputed by fixed rules before being passed to the task head.

\noindent \textbf{Auxiliary supervision and loss design.}
The remaining ablations show that the auxiliary and loss terms also contribute, but with smaller effects on the 4-class task. Removing the auxiliary PD head (A3) reduces 4-class macro F1 by $8.9$ percentage points and binary macro F1 by $10.1$ percentage points, suggesting that the PD-vs-rest supervision signal helps stabilize the clinically dominant decision boundary. Replacing the soft macro F1 with focal weighting by plain cross-entropy (A4) has a smaller effect on 4-class macro F1 ($-3.3$ percentage points), but a larger effect on the binary PD-vs-rest task ($-14.1$ percentage points). This suggests that the macro F1-aligned loss is useful under the strong class imbalance in LUMIERE, especially for the PD decision.

\noindent \textbf{Relation to the rule-based oracle.}
These ablations also help interpret the rule-based RANO oracle in Table~\ref{tab:main_comparison}. The oracle tests whether fixed thresholding of ground-truth concepts is sufficient, while B1 and B2 test what happens when the learned model loses either the expert dependency structure or the deterministic RANO derivation chain. 
Together, these results suggest that \textsc{Trace} benefits from both parts of the design, where deterministic RANO computation provides a clinically constrained reasoning path, while the learned task head can account for aspects of the expert labels that are not fully captured by segmentation-derived thresholds alone.

\subsection{Causal Effect of Concepts (CaCE) and Concept Alignment} \label{sec:concept_use_alignment}

\begin{figure}[t]
    \centering
    \includegraphics[width=0.62\textwidth]{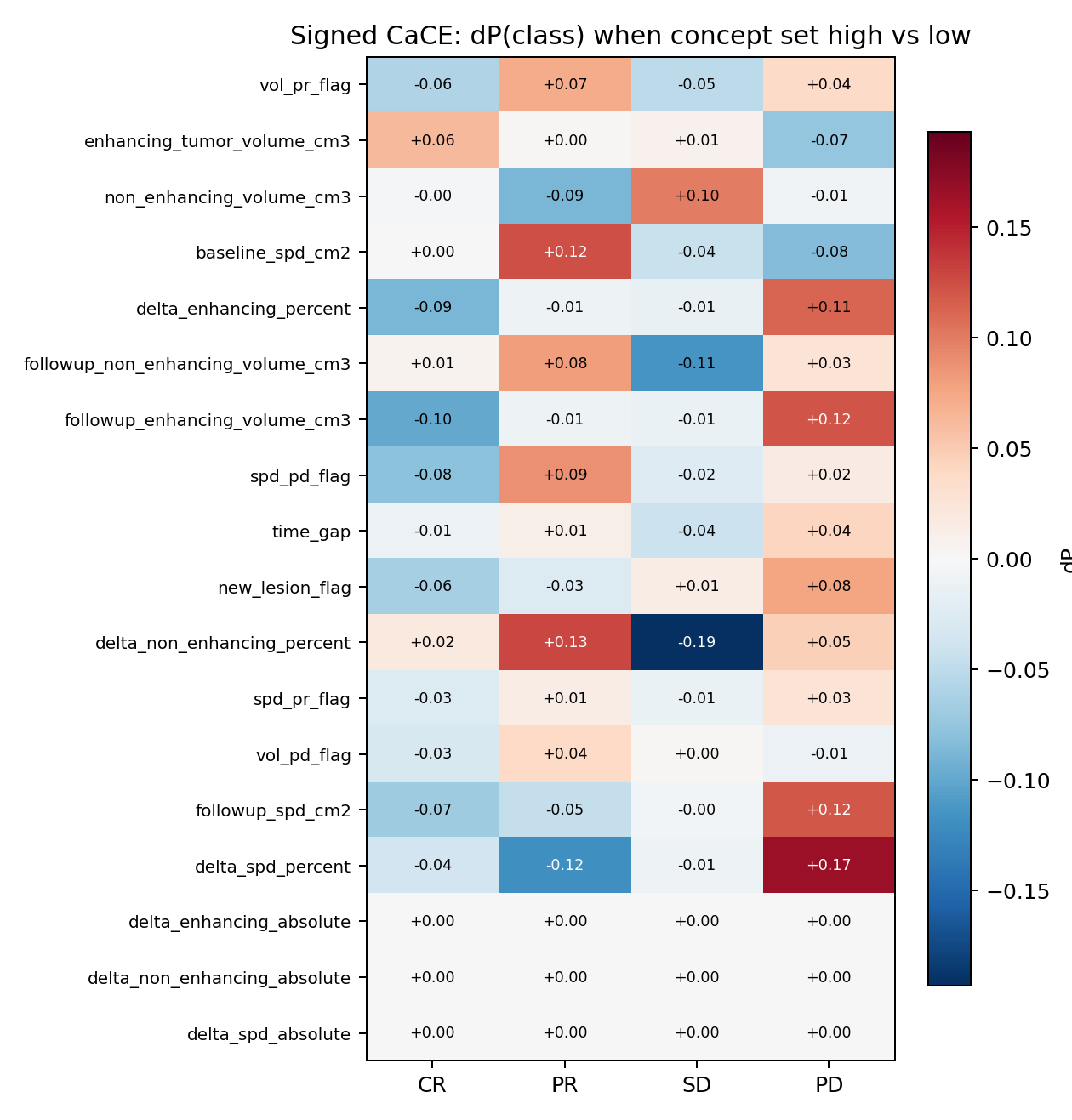}
    \caption{Signed CaCE per concept and per RANO class. Longitudinal-change concepts shift probability mass in directions consistent with RANO response categories.}
    \label{fig:cace_heatmap}
\end{figure}

We assess whether \textsc{Trace} uses the bottleneck in a clinically meaningful way through Causal Effect of Concepts total variation (CaCE-TV) \cite{goyal2019explaining,defelice2026causallyreliableconceptbottleneck}. For each concept, CaCE-TV measures the change in the 4-class output distribution when the concept is intervened from a low-value to a high-value state. Table~\ref{tab:cace} reports the top concepts by CaCE-TV, and Figure~\ref{fig:cace_heatmap} shows the signed per-class effects.

\begin{table}[ht]
\centering
\caption{Top concepts by CaCE-TV. Higher values indicate a stronger effect on the predicted RANO distribution under intervention.}
\label{tab:cace}
\begin{tabular}{llccc}
\toprule
\textbf{Rank} & \textbf{Concept}  & \textbf{CaCE} & \textbf{Std} \\
\midrule
1 & delta\_non\_enhancing\_percent & 0.195 & 0.131 \\
2 & delta\_spd\_percent & 0.170 & 0.053 \\
3 & baseline\_spd\_cm2 & 0.143 & 0.048 \\
4 & followup\_spd\_cm2 & 0.129 & 0.030 \\
5 & followup\_enhancing\_volume\_cm3 & 0.127 & 0.015 \\
6 & delta\_enhancing\_percent & 0.116 & 0.012 \\
7 & followup\_non\_enhancing\_volume\_cm3 & 0.115 & 0.084 \\
8 & spd\_pd\_flag & 0.111 & 0.059 \\
9 & vol\_pr\_flag & 0.111 & 0.070 \\
10 & non\_enhancing\_volume\_cm3 & 0.108 & 0.053 \\
11 & new\_lesion\_flag & 0.091 & 0.024 \\
12 & enhancing\_tumor\_volume\_cm3 & 0.081 & 0.024 \\
13 & time\_gap & 0.053 & 0.032 \\
14 & vol\_pd\_flag & 0.053 & 0.026 \\
15 & spd\_pr\_flag & 0.049 & 0.014 \\
16 & delta\_enhancing\_absolute & 0.000 & 0.000 \\
17 & delta\_non\_enhancing\_absolute & 0.000 & 0.000 \\
18 & delta\_spd\_absolute & 0.000 & 0.000 \\
\bottomrule
\end{tabular}
\end{table}

The CaCE results show two useful patterns. First, concept influence is distributed across the bottleneck rather than concentrated in one variable. Ten of the eighteen concepts exceed a CaCE-TV value of $0.1$, with a mean CaCE-TV of $0.110$ across non-trivial concepts (excluding the three concepts with CaCE-TV of zero). Second, the highest-effect concepts are mainly RANO-relevant change and size variables, including percentage change in non-enhancing tumor, percentage change in SPD, follow-up enhancing volume, baseline SPD, and percentage change in enhancing tumor. Their signed effects in Figure~\ref{fig:cace_heatmap} generally move probability mass in directions expected from the RANO criteria.

The main exception is \texttt{time\_gap}, which ranks thirteenth by CaCE-TV ($0.069$) but remains noteworthy as the highest-ranked non-imaging metadata concept. This suggests that scan timing carries dataset-level signal in LUMIERE. We keep \texttt{time\_gap} as a passthrough concept because scan interval is clinically available metadata. Notably, three absolute-change concepts (\texttt{delta\_enhancing\_absolute}, \texttt{delta\_non\_enhancing\_absolute}, \texttt{delta\_spd\_absolute}) yield a CaCE-TV of exactly zero, indicating that the bottleneck has learned to rely on percentage-change and SPD-derived representations rather than raw absolute differences.

\paragraph{Concept alignment.} 
We also evaluate concept alignment between predicted and target concepts. The full Concept-Alignment Score table is reported in Appendix~\ref{app:concept_alignment_details}. In summary, most RANO threshold flags are recovered with high binary accuracy, while \texttt{spd\_pd\_flag} is less reliable. Follow-up tumor measurements are also predicted more accurately than baseline measurements, suggesting that the baseline branch of the siamese encoder is a useful target for future improvement.

\subsection{Intervention Behavior}
\label{sec:hasan_intervention_curve}

\begin{figure}[ht]
\centering
\includegraphics[width=\linewidth]{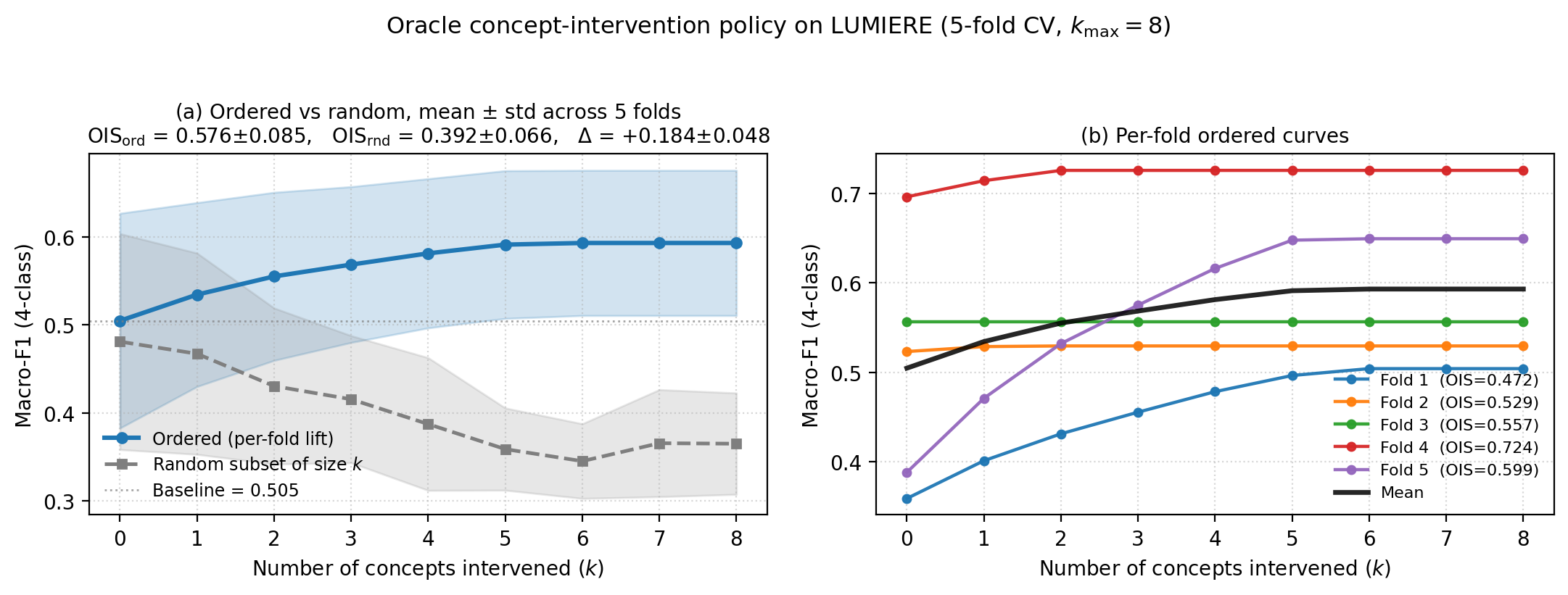}
\caption{Oracle concept-intervention policy on LUMIERE
(5-fold CV, $k_{\max}{=}8$).
\textbf{(a)} Mean 4-class macro-F1 across folds when the top-$k$ concepts are
replaced with ground-truth values, ranked by per-fold single-concept lift
(blue, ordered) or chosen at random (grey, mean over $12$ random subsets).
Shaded bands are $\pm 1$ std across folds. The ordered policy improves
macro-F1 monotonically and saturates around $k{=}5$, while the random policy
degrades the prediction once $k$ exceeds two.
\textbf{(b)} Per-fold ordered curves. Folds with a low baseline (F1, F5) gain
the most from oracle correction, while fold~4 starts at $0.696$ and saturates
within two interventions.
$\mathrm{OIS}_{\mathrm{ord}}=0.576\pm0.085$ vs
$\mathrm{OIS}_{\mathrm{rnd}}=0.392\pm0.066$, gap $+0.184\pm0.048$.}
\label{fig:ois_ordered_vs_random}
\end{figure}

We probe the bottleneck causally by replacing predicted concept values with ground-truth values and measuring the change in 4-class macro-F1. Figure~\ref{fig:ois_ordered_vs_random} reports the Ordered Intervention Score (OIS), defined as the average macro-F1 over $k=1,\ldots,8$ oracle concept replacements, computed per fold and aggregated across the five LUMIERE folds. 
The \emph{ordered} policy ranks each concept by its measured single-concept contribution in that fold; the \emph{random} policy reveals random subsets of size~$k$, averaged over $12$ random subsets per~$k$.

Across folds, $\mathrm{OIS}_{\mathrm{ord}}=0.576\pm0.085$ versus $\mathrm{OIS}_{\mathrm{rnd}}=0.392\pm0.066$, gives a per-fold gap of $+0.184\pm0.048$ macro-F1. 
The ordered curve in Figure~\ref{fig:ois_ordered_vs_random}~(a) rises sharply over the first $\sim$ five concepts and then saturates, indicating that a small set of high-impact concepts carries most of the post-intervention performance; this set overlaps with the RANO drivers identified by CaCE in Section~\ref{sec:concept_use_alignment}. 
The random curve, in contrast, drops \emph{below} the no-intervention baseline once more than two concepts are corrected, because partial corrections of correlated concepts propagated through the deterministic RANO chain can introduce inconsistencies between root measurements, percentage deltas, and threshold flags.

The per-fold view in Figure~\ref{fig:ois_ordered_vs_random}~(b) shows that the ordering benefit is consistent across folds, where folds with a low no-intervention baseline (Fold~1: $0.359$, Fold~5: $0.388$) gain the most from oracle correction ($+14$ and $+26$ percentage points respectively), while Fold~4 (i.e., already at macro-F1\,$=0.696$ at $k=0$) saturates within two interventions and provides a fold-level ceiling that bounds the achievable improvement. 
Together, these two observations support the role of intervention-consistency training, in which the bottleneck responds to ground-truth concept correction and distinguishes between informative and uninformative concepts, reflecting the kind of selective concept refinement expected in real-world clinician annotation and review.

\subsection{Patient-Level Walkthrough}\label{sec:patient_examples}

\begin{figure}[ht]
\centering
\includegraphics[width=\linewidth]{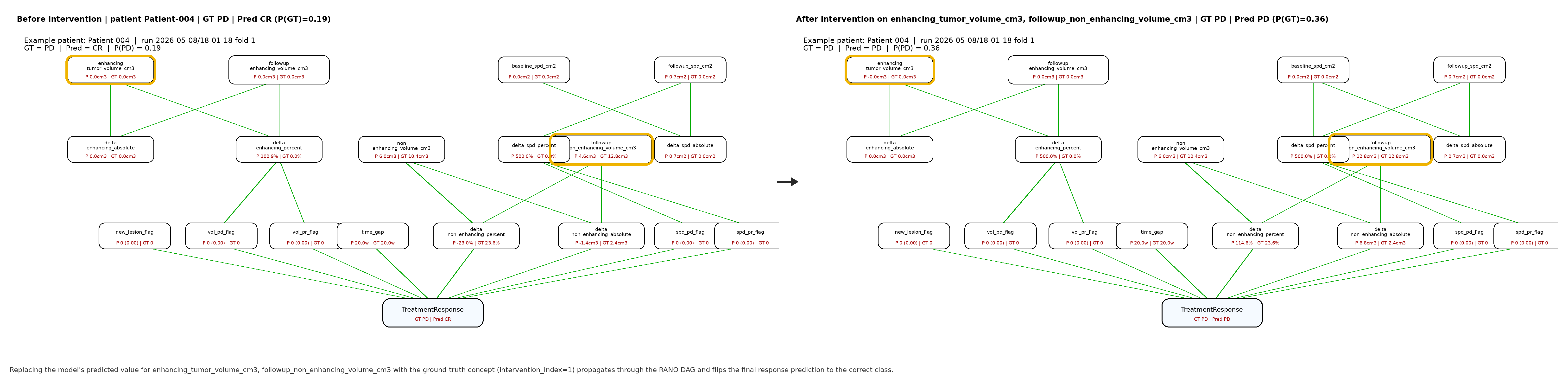}
\caption{Concept-intervention walkthrough on a misclassified case (Patient-004, validation fold~1). \textbf{Left:} default forward pass; the model predicts CR while the ground-truth label is PD. \textbf{Right:} same input with $\mathrm{intervention\_index}=1$ on \texttt{enhancing\_tumor\_volume\_cm3} and \texttt{followup\_non\_enhancing\_volume\_cm3} (yellow outlines). The corrected concept values propagate deterministically along the RANO chain ($\Delta\%\to$flags$\to$response) and the prediction flips to the correct PD class. No retraining or gradient step is performed.}
\label{fig:intervention_example}
\end{figure}

Figure~\ref{fig:intervention_example} shows a representative misclassified case (Patient-004, validation fold~1). 
In the default forward pass, \textsc{Trace} predicts CR for a true PD case, with visible errors in the volumetric root concepts, including an under-estimate of \texttt{enhancing\_tumor\_volume\_cm3}. We then correct two root concepts, \texttt{enhancing\_tumor\_volume\_cm3} and \texttt{followup\_non\_enhancing\_volume\_cm3}, both highlighted in the figure. The deterministic RANO quantities are recomputed from these corrected values, including the percentage change concepts and threshold flags, and the prediction changes from CR to PD. This example illustrates the intended use of our proposed structured bottleneck, where concept-level corrections can be propagated through the RANO-aligned graph and can change the downstream response prediction without retraining.

\section{Conclusion, Limitations, and Future Work} \label{sec:conclusion_limitations_future_work}

\paragraph{Conclusion.}
This work introduced \textsc{Trace}, an exploratory RANO-aligned concept bottleneck model for longitudinal GBM response assessment. 
This work utilized RANO-based response assessment as a structured concept reasoning problem rather than a direct image-to-label prediction task.
\textsc{Trace} estimates root tumor measurements from paired MRI, computes downstream RANO quantities deterministically, and allows concept-level corrections to propagate through the response pathway. On LUMIERE, \textsc{Trace} improved over the concept-based baseline on 4-class RANO classification and performed strongly on the PD-vs-rest decision, while preserving an inspectable bottleneck of clinical concepts. The ablation and intervention results further suggest that this behavior depends on both the expert RANO graph and intervention-consistency training.

\paragraph{Limitations.}
These results should be interpreted with several limitations in mind. First, CR and PR remain difficult because they are represented by a few cases across several validation folds, increasing fold-to-fold variation in per-class F1. Second, \texttt{time\_gap} has a strong effect on the output, suggesting that scan interval captures dataset-level signal in LUMIERE; this concept should therefore be examined carefully in external validation and may require sensitivity analysis or CaCE-based filtering. Third, concept alignment remains weaker for some baseline measurements and for the SPD-derived PD flag, suggesting that root concept estimation remains an issue. 
Finally, all experiments are based on a single dataset, i.e., LUMIERE, so broader validation on more protocol-aligned longitudinal RANO datasets is needed before drawing stronger conclusions about generalization.

\paragraph{Future work.}
Future work should prioritize validation on more datasets, stronger root concept estimation, and better handling of rare response classes. In particular, improving the estimation of baseline tumor measurements and SPD-derived progression flags may strengthen the reliability of the structured bottleneck. More broadly, the same design may extend to other longitudinal medical imaging tasks where decisions depend on explicit clinical criteria, temporal change, and human-verifiable intermediate measurements.

\section{Acknowledgments}
This work was partially funded by the Federal Ministry of Research, Technology and Space (BMFTR) under grant number 16IW23002 (No-IDLE), grant number 16IW24006 (NoIDLEChatGPT), and grant number 16IW26002 (Amenable); the Lower Saxony Ministry of Science and Culture (MWK) in the zukunft.niedersachsen program; and the Endowed Chair of AAI at the University of Oldenburg.

\section*{Declaration on Generative AI Usage}
During the preparation of this manuscript, the authors used Claude, ChatGPT, and Gemini to assist with language-related tasks, such as grammar and spelling correction, as well as to improve clarity and writing style. After using these tools, the authors carefully reviewed and edited the content as needed and take full responsibility for the final content of the publication.

\bibliography{references}

\newpage
\appendix
\section{Additional Implementation Details}\label{app:implementation_details}

\subsection{Clinical Background on SPD}
\label{app:spd_background}

The Sum of Products of Diameters (SPD) is defined as the product of the two largest perpendicular tumor diameters measured on a single imaging slice. It is retained in RANO-based response assessment because the response thresholds are historically defined using this 2D measurement. We therefore include SPD alongside volume so that the model can represent both volumetric change and the SPD-based thresholds used in the RANO criteria.

\subsection{ConceptBlock Details}
\label{app:concept_block_details}

The main model uses ConceptBlocks only for continuous image-derived root concepts. Each root concept $c_k$ is predicted from the shared image representation $\mathbf{X}_{\mathrm{enc}}$ using a concept-specific MLP:
\begin{equation}
  \mathbf{e}_k =
  \mathrm{MLP}_k(\mathbf{X}_{\mathrm{enc}})
  \in \mathbb{R}^{d_c},
\end{equation}
where $d_c$ is the concept hidden size. The scalar concept prediction is then computed as
\begin{equation}
  \hat{c}_k =
  \mathbf{w}_k^{\top}\mathbf{e}_k + b_k.
\end{equation}
The prediction $\hat{c}_k$ is trained against the z-scored ground-truth concept value using mean-squared error. The embedding $\mathbf{e}_k$ is retained as the root concept representation passed to downstream prediction.

This continuous ConceptBlock setting is used for all image-derived root concepts in this work, namely $V_{t_0}$, $V_{t_1}$, $\mathrm{SPD}_{t_0}$, and $\mathrm{SPD}_{t_1}$. Derived and passthrough nodes do not use ConceptBlocks.

\subsection{Deterministic Node Details}
\label{app:deterministic_nodes}

Deterministic nodes implement RANO-derived quantities that are fixed functions of their parent concepts. These nodes do not contain trainable parameters and are excluded from the concept supervision loss.

Because root concepts are predicted in z-scored space, deterministic computations are applied after converting predictions back to the raw clinical scale. Given a normalized prediction $\hat{c}_k$ for concept $c_k$, the raw-scale prediction is
\begin{equation}
  \tilde{c}_k =
  \mathrm{expm1}(\hat{c}_k\sigma_k+\mu_k),
\end{equation}
where $\mu_k$ and $\sigma_k$ are the training-fold mean and standard deviation of concept $c_k$. The $\mathrm{expm1}$ operation is used when the corresponding concept labels were log-transformed during preprocessing.

The volume percentage change is computed as
\begin{equation}
  \widehat{\Delta V}_{\%}
  =
  100\cdot
  \mathrm{clamp}
  \left(
    \frac{\tilde{V}_{t_1}-\tilde{V}_{t_0}}
    {\tilde{V}_{t_0}+\varepsilon},
    -1,5
  \right).
\end{equation}
The small constant $\varepsilon$ avoids division by zero. The clamp operation limits extreme ratios caused by very small baseline values and keeps the derived quantity within a stable numerical range.

The volume-based PD and PR threshold flags are then computed as
\begin{align}
  \hat{b}_{\mathrm{PD}}^{\mathrm{vol}}
  &=
  \mathbf{1}
  \left[
    \widehat{\Delta V}_{\%}\geq 40
    \;\wedge\;
    \tilde{V}_{t_0}\geq 0.5
  \right], \\[6pt]
  \hat{b}_{\mathrm{PR}}^{\mathrm{vol}}
  &=
  \mathbf{1}
  \left[
    \widehat{\Delta V}_{\%}\leq -65
    \;\wedge\;
    \tilde{V}_{t_0}\geq 0.5
  \right].
\end{align}

The SPD percentage change is computed analogously:
\begin{equation}
  \widehat{\Delta \mathrm{SPD}}_{\%}
  =
  100\cdot
  \mathrm{clamp}
  \left(
    \frac{\widetilde{\mathrm{SPD}}_{t_1}-\widetilde{\mathrm{SPD}}_{t_0}}
    {\widetilde{\mathrm{SPD}}_{t_0}+\varepsilon},
    -1,5
  \right).
\end{equation}
The SPD-based PD and PR threshold flags are
\begin{align}
  \hat{b}_{\mathrm{PD}}^{\mathrm{SPD}}
  &=
  \mathbf{1}
  \left[
    \widehat{\Delta \mathrm{SPD}}_{\%}\geq 25
    \;\wedge\;
    \widetilde{\mathrm{SPD}}_{t_0}\geq 0.01
  \right], \\[6pt]
  \hat{b}_{\mathrm{PR}}^{\mathrm{SPD}}
  &=
  \mathbf{1}
  \left[
    \widehat{\Delta \mathrm{SPD}}_{\%}\leq -50
    \;\wedge\;
    \widetilde{\mathrm{SPD}}_{t_0}\geq 0.01
  \right].
\end{align}

The measurability floors follow the thresholds in Table~\ref{tab:rano_criteria}. If the baseline measurement is below the corresponding floor, the flag is not activated for that measurement type. This avoids unstable percentage changes from very small baseline values.

Hard threshold indicators near the RANO boundaries could be relaxed using a temperature-controlled sigmoid, $\sigma((u-\tau)/T)$, for smoother gradients. In practice, the hard-threshold formulation was stable, so we use it as the default.

\subsection{Passthrough Node Details}
\label{app:passthrough_nodes}

The passthrough nodes represent concepts that are clinically relevant to the final RANO response but are not predicted from the image encoder. In this work, the passthrough nodes are the scan interval $\Delta t$ and the new-lesion flag $b_{\mathrm{NL}}$.

The scan interval $\Delta t$ is the time difference between the baseline scan $t_0$ and the follow-up scan $t_1$. It is obtained from scan metadata and injected directly into the graph:
\begin{equation}
  \hat{c}_{\Delta t} = c_{\Delta t}^{*}.
\end{equation}
This avoids treating scan timing as an image-derived concept, while still allowing the task head to use temporal context when predicting the RANO response.

The new-lesion flag $b_{\mathrm{NL}}$ indicates whether a new lesion appears at follow-up. It is computed during preprocessing from the baseline and follow-up segmentation masks. In our implementation, connected components are identified in the follow-up mask and compared with the baseline mask. A component present at $t_1$ with no corresponding component at $t_0$ is treated as evidence for a new lesion:
\begin{equation}
  b_{\mathrm{NL}} =
  \mathbf{1}
  \left[
    \exists\, L_{t_1}^{(m)} \text{ such that }
    L_{t_1}^{(m)} \cap L_{t_0} = \varnothing
  \right],
\end{equation}
where $L_{t_1}^{(m)}$ denotes the $m$-th connected component at follow-up and $L_{t_0}$ denotes the union of baseline lesion components.

The value is then injected directly into the DAG:
\begin{equation}
  \hat{c}_{b_{\mathrm{NL}}} = c_{b_{\mathrm{NL}}}^{*}.
\end{equation}
Under RANO 2.0, the appearance of a new lesion supports Progressive Disease. We therefore include $b_{\mathrm{NL}}$ as an explicit graph node rather than asking the image encoder to learn this decision implicitly.

Passthrough nodes do not have ConceptBlocks, do not have trainable parameters, and do not contribute to the concept supervision loss. They enter the graph only as scalar values supplied to downstream nodes. This keeps the distinction clear between image-derived root concepts, deterministic RANO-derived concepts, and metadata supplied outside the image encoder.

\subsection{Optional Monotonicity Constraints}
\label{app:monotonicity_constraints}

RANO reasoning implies monotonic relationships between some concepts and response classes: larger positive changes in tumor volume or SPD should not decrease the evidence for PD, and larger negative changes should not decrease the evidence for PR. These relationships can be enforced as monotonicity constraints on the task head (e.g., $\partial\,\mathrm{logit}_{\mathrm{PD}}/\partial \Delta V_{\%}\geq 0$ and $\partial\,\mathrm{logit}_{\mathrm{PR}}/\partial \Delta V_{\%}\leq 0$, with the analogous constraints on $\Delta\mathrm{SPD}_{\%}$), implemented via constrained linear layers or monotonic networks. We treat this as an optional extension rather than a required component of the main model because hard monotonic constraints can be too restrictive on noisy clinical measurements, and the intervention-consistency objective already enforces the dominant directional behavior empirically.

\subsection{External Dataset Considerations}
\label{app:external_datasets}

We examined whether external longitudinal glioma datasets could be used to supplement LUMIERE or provide additional validation data. In particular, we considered UCSD-PTGBM~\citep{gagnon2025ucsd} and UCSF-ALPTDG~\citep{fields2024ucsf}. These datasets are useful resources for longitudinal post-treatment glioma MRI analysis, but they do not provide scan-pair-level expert RANO response labels aligned with the four categories used in this work, namely CR, PR, SD, and PD.

This matters because \textsc{Trace} is designed around RANO-based response assessment. The model predicts the final RANO label and structures its intermediate concepts around RANO measurements, threshold flags, and passthrough clinical metadata. Labels based on other clinical endpoints, dataset-specific categories, survival information, or post hoc thresholds are therefore not interchangeable with expert RANO response labels.

We tested whether approximate RANO labels could be reconstructed for the external datasets using segmentation-derived longitudinal volume changes. This was not suitable because RANO assessment is not determined by volume change alone. It also depends on measurable disease requirements, SPD-based thresholds, the appearance of new lesions, and assessment context. Reconstructed labels were especially ambiguous near decision boundaries between PR, SD, and PD.

For these reasons, we did not merge the external datasets with LUMIERE for model training. Doing so would have increased the number of scans, but it would also have mixed different label definitions. We therefore restrict the main experiments to LUMIERE, where the supervision target matches the RANO-based task. External validation remains an important future direction, but it requires datasets with protocol-aligned scan-pair-level RANO annotations.

\subsection{Preprocessing Details}
\label{app:preprocessing_details}

\begin{figure}[htbp]
\centering
\includegraphics[width=0.9\linewidth]{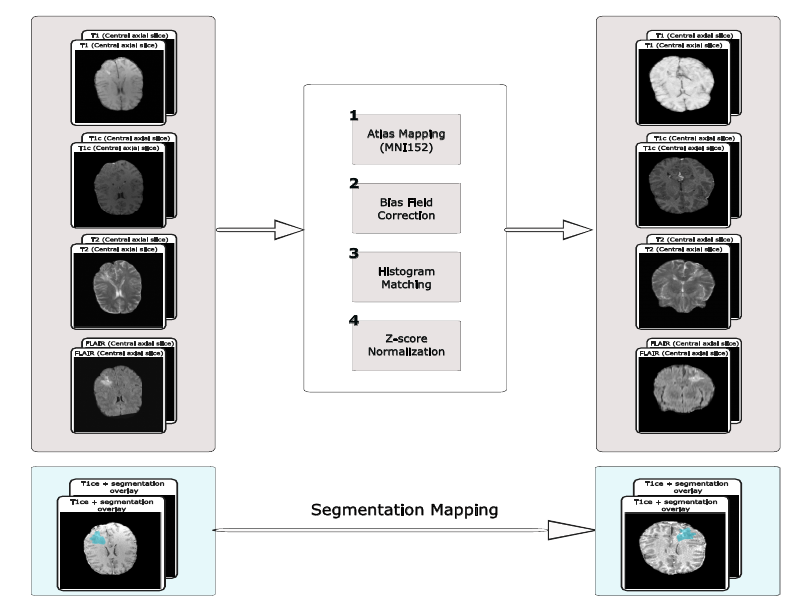}
\caption{Preprocessing pipeline applied to the raw MRI data. The four MRI modalities undergo atlas registration to MNI152 space, N4 bias field correction, histogram matching, and z-score normalization. The HD-GLIO-AUTO segmentation mask is mapped to the same template space using the transformation estimated from the corresponding MRI, with nearest-neighbor interpolation used to preserve discrete label values.}
\label{fig:preprocessing}
\end{figure}

All MRI sequences were spatially normalized before model training. The four modalities were registered to the MNI152 1\,mm template space. N4 bias field correction~\citep{tustison2010n4itk} was applied to reduce low-frequency intensity inhomogeneity. Histogram matching was then used to reduce intensity variation across scans, followed by z-score normalization.
Segmentation masks were transformed using the same image-to-template transformation estimated from the corresponding MRI. Nearest-neighbor interpolation was used to preserve discrete label values. This ensured that the MRI volumes and tumor segmentation masks remained spatially aligned after registration. The final model input was resampled to $128{\times}128{\times}128$ voxels.
Figure~\ref{fig:preprocessing} shows a summary of the 4-step preprocessing pipeline.

\subsection{Data-Efficient CBM Baseline Details}
\label{app:dcbm_details}

We adapted the Data-Efficient CBM baseline~\citep{prasse2025dcbmdataefficientvisualconcept} to the longitudinal 3D MRI setting. For each scan, tissue-density-filtered 3D patches of size $32 \times 64 \times 64$ voxels were extracted across the four MRI modalities. Tissue-density filtering was used to reduce background-dominated patches and focus the baseline on image regions more likely to contain anatomical or tumor-related information.

Each patch was encoded using a frozen M3D-LaMed-Phi-3 encoder~\citep{bai2024m3d}, producing a 3072-dimensional embedding. These embeddings were passed to a sparse autoencoder with 128 concepts and sparsity weight $\lambda=0.001$. The sparse autoencoder was trained separately within each cross-validation fold to avoid leakage between training and validation splits.

For each baseline--follow-up pair, longitudinal change was represented by subtracting the baseline concept activations from the follow-up concept activations. The resulting difference representation was used as input to an $L_2$-regularized logistic regression classifier for 4-class RANO response prediction. This setup provides a concept-based comparison in which concepts are discovered from pretrained embeddings rather than defined from RANO measurements and clinical rules.

\subsection{Training Hyperparameters}
\label{app:training_hyperparams}

Table~\ref{tab:hyperparams} summarizes the main training hyperparameters used for \textsc{Trace} variants.

\begin{table}[htbp]
\centering
\caption{Training hyperparameters for \textsc{Trace} variants.}
\label{tab:hyperparams}
\setlength{\tabcolsep}{4pt}
\begin{tabular}{ll}
\toprule
\textbf{Hyperparameter} & \textbf{Value} \\
\midrule
Optimizer & AdamW \\
Learning rate & $10^{-3}$ \\
$L_2$ weight decay & $0.01$ \\
Batch size & $16$ \\
Input dimensions & $128{\times}128{\times}128$ voxels \\
Max epochs & $100$ \\
LR scheduler & ReduceLROnPlateau \\
 & factor $0.5$, patience $10$ \\
Early stopping patience & $20$ epochs, validation macro F1 \\
Gradient clipping & $\|\nabla\|_2 \leq 1.0$ \\
Dropout & $p = 0.3$ \\
Task loss & Soft macro F1 with focal weighting \\
Auxiliary loss weight $\alpha_{\mathrm{aux}}$ & $0.1$ \\
Intervention consistency weight $\beta_{\mathrm{cons}}$ & $0.5$ \\
Concept loss weight $\lambda_c$ & $0.8$ \\
Intervention probability $p_{\mathrm{int}}$ & annealed from $0.3$ to $0.8$ \\
Annealing period & first $15$ epochs \\
$L_1$ penalty & $10^{-4}$ on the first layer of the task head \\
Encoder freezing & epoch $50$ \\
\bottomrule
\end{tabular}
\end{table}

\subsection{Ordered vs Random Intervention Policy Ablation}
\label{app:intervention_policy_ablation}

To test whether concept ranking is meaningful for interventions, we compared an ordered policy (top-$k$ concepts ranked by per-concept intervention lift) against a random-$k$ policy. Table~\ref{tab:intervention_policy_ablation_appendix} reports the 4-class macro F1 results from the run in \texttt{paper\_figures/intervention\_policy\_ablation\_k6\_r6.csv}. The ordered policy consistently outperforms random interventions for $k \geq 2$, and the gap increases with larger $k$, indicating that the ranking captures actionable concept importance.

\begin{table}[ht]
\centering
\caption{Ordered vs random intervention policy ablation (4-class macro F1, mean $\pm$ std).}
\label{tab:intervention_policy_ablation_appendix}
\setlength{\tabcolsep}{6pt}
\small
\begin{tabular}{cccc}
\toprule
\textbf{$k$} & \textbf{Ordered policy} & \textbf{Random policy} & \textbf{Gap (Ordered - Random)} \\
\midrule
0 & $0.481 \pm 0.123$ & $0.481 \pm 0.123$ & $0.000 \pm 0.000$ \\
1 & $0.481 \pm 0.123$ & $0.482 \pm 0.119$ & $-0.001 \pm 0.010$ \\
2 & $0.481 \pm 0.123$ & $0.437 \pm 0.096$ & $0.044 \pm 0.027$ \\
3 & $0.481 \pm 0.123$ & $0.411 \pm 0.071$ & $0.070 \pm 0.065$ \\
4 & $0.481 \pm 0.123$ & $0.411 \pm 0.067$ & $0.070 \pm 0.077$ \\
5 & $0.481 \pm 0.123$ & $0.399 \pm 0.070$ & $0.082 \pm 0.073$ \\
6 & $0.481 \pm 0.123$ & $0.392 \pm 0.074$ & $0.089 \pm 0.088$ \\
\bottomrule
\end{tabular}
\end{table}

\subsection{Fold-Level Metrics}
\label{app:fold_metrics}

Table~\ref{tab:fold_metrics_appendix} reports the per-fold metrics for \textsc{Trace}. These values support the aggregate results in Section~\ref{sec:results_discussion}. Variation across folds is mainly driven by the minority CR and PR classes, while the PD-vs-rest binary task is more stable.

\begin{table}[ht]
\centering
\caption{Per-fold and aggregate metrics for \textsc{Trace}.}
\label{tab:fold_metrics_appendix}
\setlength{\tabcolsep}{4pt}
\small
\begin{tabular}{lcccc}
\toprule
\textbf{Fold} & \textbf{4-class macro F1} & \textbf{4-class weighted acc} & \textbf{Binary macro F1} & \textbf{Binary weighted acc} \\
\midrule
F1 & 0.359 & 0.402 & 0.616 & 0.621 \\
F2 & 0.523 & 0.464 & 0.656 & 0.662 \\
F3 & 0.418 & 0.616 & 0.778 & 0.841 \\
F4 & 0.696 & 0.768 & 0.857 & 0.891 \\
F5 & 0.388 & 0.473 & 0.635 & 0.651 \\
\midrule
\textbf{Mean $\pm$ Std} 
& \textbf{$0.4769 \pm 0.1229$} 
& \textbf{$0.5447 \pm 0.1318$} 
& \textbf{$0.7085 \pm 0.0935$} 
& \textbf{$0.7329 \pm 0.1104$} \\
\bottomrule
\end{tabular}
\end{table}

\subsection{Per-Class Metrics}
\label{app:per_class_metrics}

Table~\ref{tab:per_class_metrics_appendix} reports the per-class F1 values across folds. The table complements Figure~\ref{fig:perclass_f1} in the main text.

\begin{table}[ht]
\centering
\caption{Per-class F1 across folds. The mean column aggregates all five folds; the displayed fold columns show folds where each class has non-trivial support.}
\label{tab:per_class_metrics_appendix}
\setlength{\tabcolsep}{4pt}
\small
\begin{tabular}{llccccc}
\toprule
\textbf{Task} & \textbf{Class} & \textbf{Mean $\pm$ Std} & \textbf{F1} & \textbf{F2} & \textbf{F3} & \textbf{F4} \\
\midrule
4-class & CR     & $0.405 \pm 0.295$ & 0.000 & 0.400 & 0.222 & 0.875 \\
4-class & PR     & $0.270 \pm 0.235$ & 0.316 & 0.571 & 0.000 & 0.462 \\
4-class & SD     & $0.477 \pm 0.104$ & 0.489 & 0.359 & 0.606 & 0.571 \\
4-class & PD     & $0.756 \pm 0.095$ & 0.632 & 0.763 & 0.842 & 0.877 \\
Binary  & non-PD & $0.661 \pm 0.103$ & 0.600 & 0.549 & 0.714 & 0.837 \\
Binary  & PD     & $0.756 \pm 0.095$ & 0.632 & 0.763 & 0.842 & 0.877 \\
\bottomrule
\end{tabular}
\end{table}

\subsection{Rule-Based RANO Oracle}
\label{app:rule_based_rano_reference}

The rule-based RANO oracle applies fixed RANO threshold rules directly to the ground-truth concept vector for each validation pair. A case is assigned to PD if a new lesion is present, if the enhancing-volume change satisfies the PD threshold, or if the SPD change satisfies the PD threshold. PR and CR are assigned using the corresponding RANO threshold rules, with SD used when no progression or response rule is activated.

This reference is included to test whether fixed thresholding of the available concept values is sufficient for the LUMIERE response labels. It reaches $0.346 \pm 0.038$ 4-class macro F1 and $0.681 \pm 0.058$ binary macro F1. The result is below \textsc{Trace}, indicating that the expert labels in LUMIERE are not fully recovered by segmentation-derived threshold rules alone. This is plausible because clinical response assessment can also reflect non-measurable disease, assessment context, and other clinical factors not represented in the segmentation-derived concepts.

\subsection{Additional Concept Causal Effects Details}
\label{app:concept_effects_details}

\begin{figure}[ht]
    \centering
    \includegraphics[width=0.62\textwidth]{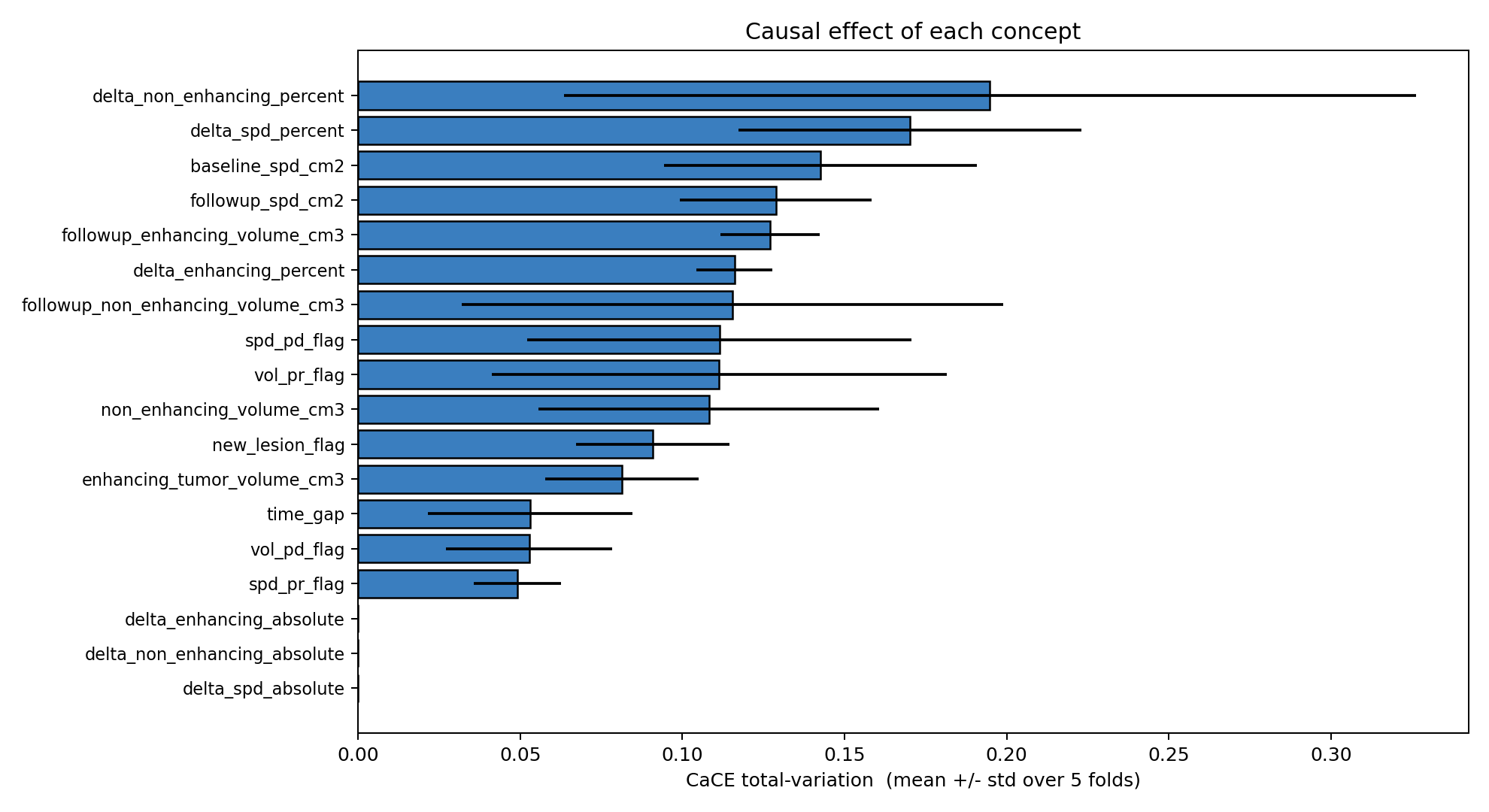}
    \caption{Top concepts by CaCE-TV. Concept influence is distributed across multiple RANO-related variables rather than concentrated in a single concept.}
    \label{fig:cace_bar}
\end{figure}

Figure~\ref{fig:cace_bar} summarizes the overall magnitude of concept influence by ranking concepts according to their CaCE-TV scores, independent of the direction of their class-specific effects.

\subsection{Concept-Alignment Details}
\label{app:concept_alignment_details}

Table~\ref{tab:concept_alignment_appendix} reports concept-alignment results across the five folds. Binary accuracy is used for flag concepts, and z-scored MAE is used for continuous concepts, where lower values indicate better alignment.
Volume-based threshold flags are recovered with high binary accuracy (\texttt{vol\_pd\_flag}: $0.990 \pm 0.015$; \texttt{vol\_pr\_flag}: $0.968 \pm 0.029$), suggesting that the bottleneck reliably captures the volumetric progression and response signals. 
However, \texttt{spd\_pd\_flag} achieves only $0.399 \pm 0.085$ binary accuracy, which indicates that the SPD-based progression pathway is not reliably predicted, and we assume that this suggests the model relies more on volume-based concepts than the RANO graph implies.

Among continuous concepts, follow-up measurements are consistently predicted more accurately than their baseline counterparts, e.g., \texttt{followup\_enhancing\_volume\_cm3} achieves a z-scored MAE of $0.295 \pm 0.107$ compared to $0.640 \pm 0.505$ for \texttt{enhancing\_tumor\_volume\_cm3}, which may partly reflect the segmentation mask dependency discussed in the main limitations, since the encoder receives masks as input while supervision targets are derived from those same masks. 
SPD-based continuous concepts (\texttt{delta\_spd\_percent}: $0.840 \pm 0.072$; \texttt{baseline\_spd\_cm2}: $0.856 \pm 0.308$) show the weakest alignment overall, pointing to SPD estimation from volumetric 3D features as a key target for future improvement. We plan to investigate these assumptions further in future work using additional datasets. 

\begin{table}
\centering
\caption{Concept-alignment summary across five folds. Binary accuracy is reported for flag concepts; z-scored MAE is reported for continuous concepts.}
\label{tab:concept_alignment_appendix}
\setlength{\tabcolsep}{6pt}
\small
\begin{tabular}{lcl}
\toprule
\textbf{Concept} & \textbf{Score $\pm$ Std} & \textbf{Metric} \\
\midrule
\texttt{new\_lesion\_flag}                    & $1.000 \pm 0.000$ & binary accuracy \\
\texttt{vol\_pd\_flag}                        & $0.990 \pm 0.015$ & binary accuracy \\
\texttt{vol\_pr\_flag}                        & $0.968 \pm 0.029$ & binary accuracy \\
\texttt{spd\_pr\_flag}                        & $0.864 \pm 0.040$ & binary accuracy \\
\texttt{spd\_pd\_flag}                        & $0.399 \pm 0.085$ & binary accuracy \\
\midrule
\texttt{followup\_enhancing\_volume\_cm3}      & $0.295 \pm 0.107$ & z-score MAE \\
\texttt{followup\_spd\_cm2}                    & $0.434 \pm 0.070$ & z-score MAE \\
\texttt{followup\_non\_enhancing\_volume\_cm3} & $0.552 \pm 0.179$ & z-score MAE \\
\texttt{enhancing\_tumor\_volume\_cm3}         & $0.640 \pm 0.505$ & z-score MAE \\
\texttt{delta\_non\_enhancing\_percent}        & $0.688 \pm 0.217$ & z-score MAE \\
\texttt{delta\_enhancing\_percent}             & $0.745 \pm 0.060$ & z-score MAE \\
\texttt{delta\_spd\_percent}                   & $0.840 \pm 0.072$ & z-score MAE \\
\texttt{baseline\_spd\_cm2}                    & $0.856 \pm 0.308$ & z-score MAE \\
\texttt{non\_enhancing\_volume\_cm3}           & $0.895 \pm 0.112$ & z-score MAE \\
\bottomrule
\end{tabular}
\end{table}

\end{document}